\pdfoutput=1

\documentclass[11pt]{article}

\usepackage{ACL2025}

\usepackage{times}
\usepackage{latexsym}
\usepackage{graphicx} 
\usepackage{amsmath} 
\usepackage{algpseudocode}
\usepackage{booktabs}
\usepackage{float}
\usepackage{multirow}
\usepackage{enumitem}
\usepackage[utf8]{inputenc}

\usepackage{enumitem}

\usepackage[vlined, ruled]{algorithm2e}
\usepackage{colortbl}

\usepackage{xcolor}
\usepackage{tcolorbox}

\usepackage{arydshln}

\definecolor{originalq}{RGB}{227, 242, 253}  
\definecolor{privacyissue}{RGB}{255, 235, 238}  
\definecolor{finalq}{RGB}{232, 245, 233}  
\definecolor{evidence}{RGB}{255, 243, 224}  
\definecolor{filtered}{RGB}{243, 229, 245}  
\definecolor{rag}{RGB}{224, 247, 250}  
\definecolor{response}{RGB}{252, 228, 236}  
\definecolor{eval}{RGB}{241, 248, 233}  
\definecolor{table1blue}{RGB}{146,172,209}

\newtcolorbox{originalquestion}[1][]{
    colback=originalq,
    title=Original Question,
    #1
}
\newtcolorbox{Benchmark Question}[1][]{
    colback=benchq,
    title=Benchmark Question,
    #1
}
\newtcolorbox{privacyissuequestion}[1][]{
    colback=privacyissue,
    title=Modified Question with Privacy Issue,
    #1
}

\newtcolorbox{finalquestion}[1][]{
    colback=finalq,
    title=Final Question (Privacy Issue Removed),
    #1
}

\newtcolorbox{evidencegen}[1][]{
    colback=evidence,
    title=Evidence Generation,
    #1
}

\newtcolorbox{evidencefilt}[1][]{
    colback=filtered,
    title=Evidence Filtering,
    #1
}

\newtcolorbox{ragrel}[1][]{
    colback=rag,
    title=RAG Relationship Generation,
    #1
}

\newtcolorbox{ragfilt}[1][]{
    colback=filtered,
    title=RAG Filtering,
    #1
}

\newtcolorbox{responsecol}[1][]{
    colback=response,
    title=Response Collection,
    #1
}

\newtcolorbox{responseeval}[1][]{
    colback=eval,
    title=Response Evaluation,
    #1
}

\definecolor{lightgray}{gray}{0.9} 
\definecolor{lightblue}{RGB}{220,230,241}

\usepackage[T1]{fontenc}

\usepackage[utf8]{inputenc}

\usepackage{microtype}

\usepackage{inconsolata}

\newcommand{\name}{\texttt{DRAG}}

\title{\texttt{DRAG}: Distilling RAG for SLMs from LLMs to Transfer Knowledge and Mitigate Hallucination via Evidence and Graph-based Distillation}

\author{Jennifer Chen\thanks{~~Equal contribution. Work done while Jennifer and Hassaan visiting VILA Lab at MBZUAI, supervised by Zhiqiang Shen. $^{\S}$Corresponding author.}~$^{,\ddag,\dag}$, Aidar Myrzakhan$^{*,\ddag}$, Yaxin Luo$^{\ddag}$, Hassaan Muhammad Khan$^{\ddag,\natural}$\\ \bf Sondos Mahmoud Bsharat$^{\ddag}$, Zhiqiang Shen$^{\ddag,\S}$ \\
  $^{\ddag}$VILA Lab, Mohamed bin Zayed University of AI \\
  $^{\dag}$McGill University \\
  $^{\natural}$National University of Science and Technology \\
  }

\begin{document}
\maketitle
\begin{abstract}
Retrieval-Augmented Generation (RAG) methods have proven highly effective for tasks requiring factual consistency and robust knowledge retrieval. However, large-scale RAG systems consume significant computational resources and are prone to generating ``hallucinated'' content from Humans\footnote{Human incorrect answers can pollute RAG's database.}. In this work, we introduce \name{}, a novel framework for distilling RAG knowledge from large-scale Language Models (LLMs) into small LMs (SLMs). Our approach leverages evidence- and knowledge graph–based distillation, ensuring that the distilled model retains critical factual knowledge while significantly reducing model size and computational cost. By aligning the smaller model's predictions with a structured knowledge graph and ranked evidence, \name{} effectively mitigates hallucinations and improves factual accuracy. We further present a case demonstrating how our framework mitigates user privacy risks and introduce a corresponding benchmark. Experimental evaluations on multiple benchmarks demonstrate that our method outperforms the prior competitive RAG methods like MiniRAG for SLMs by up to 27.7\% using the same models, preserving high-level efficiency and reliability. With \name{}, we provide a practical and resource-efficient roadmap to deploying enhanced retrieval and generation capabilities in small-sized LLMs. Code is available at \url{https://github.com/VILA-Lab/DRAG}.
\end{abstract}

\section{Introduction}

The development of retrieval-augmented generation (RAG) frameworks has significantly advanced the capabilities of large language models (LLMs) by integrating external knowledge retrieval with generative capabilities. RAG models allow for dynamic retrieval of evidence, enhancing both factual accuracy and contextual relevance. However, these frameworks are computationally expensive by maintaining an up-to-date large-scale knowledge base and are primarily designed for large-scale LLMs, making them impractical for smaller LLMs deployed on resource-constrained environments. Furthermore, the hallucination problem, where the model generates plausible-sounding but factually incorrect information, remains a critical challenge even in advanced RAG systems. Addressing these issues is crucial for the effective utilization of LLMs in real-world applications.

In this work, we introduce \name{} ({\bf Distilling RAG}), a novel approach aimed at transferring the knowledge and capabilities of large models to smaller LLMs while simultaneously mitigating hallucination through evidence-based distillation. Our method is motivated by the need to make RAG frameworks more accessible and efficient for smaller models without compromising their ability to retrieve and generate accurate information. By leveraging the retrieval process as a core component of distillation, \name{} provides a structured mechanism to teach smaller LLMs how to ground their outputs in external evidence.

\begin{figure*}[t]
    \centering
    \includegraphics[width=0.95\linewidth]{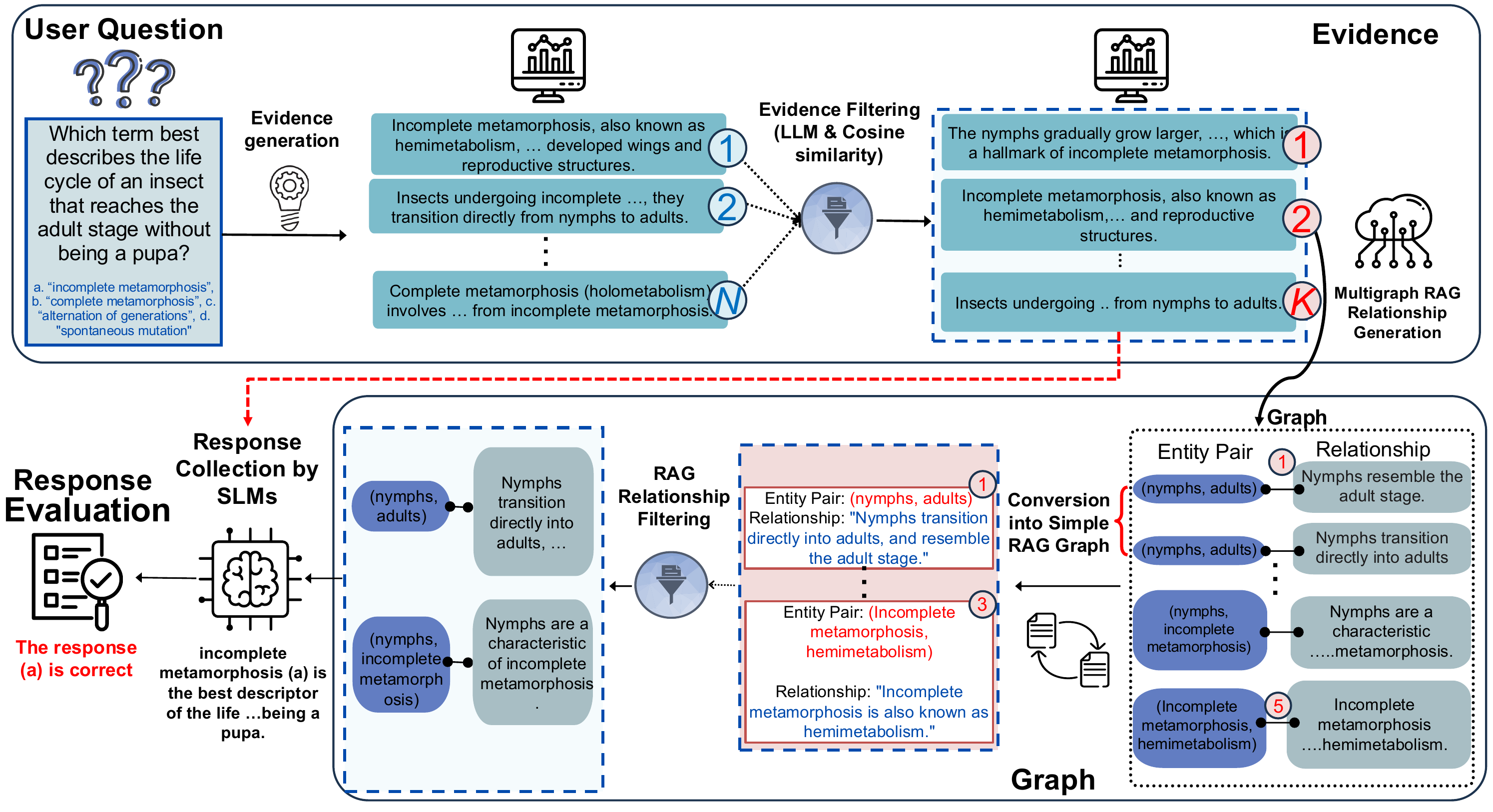}
    \vspace{-0.1in}
    \caption{{\bf Framework Overview of Our Evidence- and Graph-based RAG Distillation.} Given a user query (top-left), the approach first retrieves and filters evidence by collecting relevant text snippets. Then, these references are fed into relationship filtering and ranking using an LLM and cosine similarity to yield high-quality ordered references. The resulting multigraph RAG  structure is then converted into a simplified RAG graph (bottom-right), distilling crucial relationships and factual context, also extracts key entity pairs and links (e.g., ``nymphs$\rightarrow$adults''). Finally, SLMs leverage this distilled information, mitigating hallucination and transferring knowledge effectively.}
    \label{fig:methodology}
    \vspace{-0.1in}
\end{figure*}

The proposed \name{} framework employs a multi-stage paradigm as in Figure~\ref{fig:methodology}. First, it generates associated evidences and knowledge graphs based on the context to the input questions from a large RAG teacher model. Then, it distills the retrieval and generation of knowledge into a smaller student / target model. By aligning the student's retrieval and generation processes with those of the teachers, \name{} ensures that the student model can effectively mimic the evidence-driven reasoning of the teacher. We further introduce an evidence-based privacy protection mechanism to reduce privacy issues, as an additional use case of our proposed framework. To achieve this, we construct a new benchmark dataset with information leakage. Then, we let the local target model rephrase input questions before uploading them to a cloud-based large-scale teacher LLM to generate corresponding evidence and knowledge graphs. Finally, the target model utilizes these received evidence and knowledge graphs to produce more reliable and accurate answers. 

To evaluate the effectiveness of \name{}, we conduct extensive experiments across various datasets and tasks. Our results demonstrate that \name{} significantly enhances the performance of SLMs in retrieval-augmented tasks by more than 20\%, achieving results comparable to their larger teacher models. Moreover, \name{} consistently outperforms baseline RAG methods in mitigating hallucination, as evidenced by improved factual accuracy and reduced error in generated outputs. This advantage stems from the teacher LLM's ability to generate more relevant and abstract evidence and knowledge graphs, making them easier for SLMs to interpret and utilize effectively. These findings highlight the potential of \name{} to bridge the gap between LLMs and SLMs in a retrieval-augmented setting.

The contributions of this paper are threefold:

{\bf 1)} We propose a novel evidence and knowledge graph-based distillation framework for transferring RAG capabilities and mitigating hallucination from large to small-scale LLMs. 

{\bf 2)} We construct a privacy leakage benchmark and introduce a privacy mitigation mechanism based on our framework that integrates evidence consistency to demonstrate the additional advantage and strong applicability of our approach. 

{\bf 3)} We provide a comprehensive evaluation of \name{} on diverse tasks and datasets, as well as various teacher LLMs and student SLMs, showing its superior ability to balance efficiency, accuracy, and factual consistency.

In summary, by allowing SLMs to harness the strengths of RAG frameworks from a distillation scheme, \name{} opens new opportunities to deploy powerful and reliable LLMs in resource-constrained settings, offering a way for their wider adoption in real-world applications.

\section{Related Work}
\vspace{-0.05in}

 RAG frameworks have been widely explored for tasks requiring factual accuracy and enhanced knowledge retrieval. The foundational work \cite{lewis2020} introduced the RAG model, which integrates dense neural retrievers with sequence-to-sequence language models, achieving state-of-the-art performance on knowledge-intensive tasks. Subsequent research has focused on refining the retrieval mechanisms within RAG frameworks \cite{he2024g, yan2024corrective, wang2023knowledgpt}. For example, Active Retrieval Augmented Generation \cite{jiang-etal-2023-active} introduces a dynamic retrieval strategy that selects information based on the input query, thereby improving generation quality. Similarly, a unified active retrieval approach was proposed to employ multiple criteria for assessing the necessity of retrieval, optimizing the overall generation process \cite{cheng-etal-2024-unified}.

Incorporating structured knowledge into RAG has garnered significant interest in addressing hallucination and enhancing factual grounding. Graph-based methods, such as Graph RAG \cite{edge2024local}, construct an entity knowledge graph from source documents, enabling large language models to handle global questions over entire text corpora. This method enhances query-focused summarization by leveraging graph-based text indices, leading to substantial improvements in the comprehensiveness and diversity of generated answers. Therefore, other graph-based methods have been explored that utilize graph structures to improve both retrieval precision and generative coherence \cite{hu2024grag, mao2024advancing, mavromatis2024gnn}. Similarly, a framework was proposed to align retrieval processes with knowledge graph structures, improving logical consistency in generated responses \cite{ma2024think}.

Recently, ranking-based methods like LambdaMART~\cite{burges2010ranknet} with RRF~\cite{cormack2009reciprocal} enhance RAG by refining retrieval and reducing hallucinations. However, their effectiveness is limited by small context windows and reliance on synthetic data \cite{anantha2023context}. To overcome this, long-context LLMs have been integrated to handle larger retrieval units, improving both retrieval and generation performance while reducing the retriever’s workload \cite{zhu2024longragenhancingretrievalaugmentedgeneration}. This integration has shown promising gains, particularly in tasks requiring deep contextual understanding \cite{xu2023retrieval}.

Several efforts have preliminarily explored distillation techniques for RAG systems~\cite{izacard2021distillingretrievalaugmentedgeneration,jia2024bridging,bezerra2025llmquoter}. For instance, LLMQuoter~\cite{bezerra2025llmquoter} fine-tunes a model using Low-Rank Adaptation (LoRA) on a 15k-sample subset of HotpotQA and employs a {\em quote-first-then-answer} strategy. In contrast, our method is finetuning-free and more efficient.
Our approach focuses on enhancing response quality, factual consistency, and retrieval accuracy by integrating evidence and knowledge graphs.

\section{Method}

\subsection{Preliminaries}
\textbf{Naive RAG.} Let $\mathcal{X}$ denote the space of input queries and $\mathcal{Y}$ the space of possible outputs. To enhance the generation process, RAG leverages a large external corpus $\mathcal{D} = \{d_1, d_2, \ldots, d_{|\mathcal{D}|}\}$. Given an input query $\mathbf{x} \in \mathcal{X}$, the framework retrieves relevant documents $\mathbf{d}_k$ from $\mathcal{D}$ and uses these retrieved documents to condition the output generation. RAG decomposes conditional output distribution $p(\mathbf{y} \mid \mathbf{x})$. Formally, the output distribution is represented as:
\vspace{-0.05in}
\begin{equation}
    p(\mathbf{y} \mid \mathbf{x}) = \sum_{\mathbf{d} \in \mathcal{D}} p_{\theta_r}(\mathbf{d} \mid \mathbf{x}) \, p_{\theta_g}(\mathbf{y} \mid \mathbf{x}, \mathbf{d}),
\vspace{-0.05in}
\end{equation}
where $p_{\theta_r}(\mathbf{d} \mid \mathbf{x})$ is referred to as the retrieval distribution, which assigns a relevance score to each document $\mathbf{d}$ given the input query $\mathbf{x}$, and $p_{\theta_g}(\mathbf{y} \mid \mathbf{x}, \mathbf{d})$ is the generation distribution, which generates the final output $\mathbf{y}$ while attending to both the input $\mathbf{x}$ and the retrieved document $\mathbf{d}$, where $\theta$ represents the model parameters.

\noindent\textbf{Graph RAG.}
Let $\mathcal{G} = \mathcal{V}, \mathcal{E}$ be a knowledge graph, where each node $v \in \mathcal{V}$ denotes an entity (e.g., a concept or object) and each edge $(v_i,v_j) \in \mathcal{E}$ captures a relationship between entities $v_i$ and $v_j$. Rather than retrieving individual documents $\mathbf{d}$ from a corpus, Graph RAG seeks relevant \emph{subgraphs} $\mathbf{z}$ within $\mathcal{G}$ to provide structured context for generation.
A subgraph $\mathbf{z}$ can be formed by selecting a subset of nodes (and their induced edges) that are topically or semantically related to the input query $\mathbf{x}$. Formally, the framework factorizes the conditional distribution $p(\mathbf{y} \mid \mathbf{x})$ as:
\vspace{-0.05in}
\begin{equation}
    p(\mathbf{y} \mid \mathbf{x}) = \sum_{\mathbf{z} \subseteq \mathcal{G}} p_{\theta_r}(\mathbf{z} \mid \mathbf{x}) \, p_{\theta_g}(\mathbf{y} \mid \mathbf{x}, \mathbf{z}),
\vspace{-0.05in}
\end{equation}
where $p_{\theta_r}(\mathbf{z}\!\mid\!\mathbf{x})$ is the retrieval distribution that assigns a relevance score to each subgraph $\mathbf{z}$, and $p_{\theta_g}(\mathbf{y}\!\!\mid\!\!\mathbf{x}, \mathbf{z})$ is the generation distribution conditioned on both the input $\mathbf{x}$ and retrieved subgraph.

\begin{algorithm*}[t!]
\small
\LinesNumberedHidden
\caption{\textbf{\texttt{DRAG} Framework}}
\label{alg:mrag}

\begin{minipage}[t]{0.4\textwidth}

\textbf{Step 1: Evidence Generation}
  
    \KwIn{ A question $q$; 
    a large-scale LLM $\mathcal{M}_\text{large}$; 
    a small-scale LLM $\mathcal{M}_\text{small}$; 
    number of evidences to generate $N$; 
    number of top relationships $K$
    }
    \ForEach{question $q$}{
      \nlset{1} Prompt $\mathcal{M}_\text{large}$ to generate $N$ textual evidences relevant to $q$. 
      \begingroup
      \setlength{\abovedisplayskip}{2pt}
      \setlength{\belowdisplayskip}{2pt}
      \[
          \mathcal{D} = \{ d_1, d_2, \dots, d_{N} \}
      \]
      \endgroup
  }
\end{minipage}
\hfill
\begin{minipage}[t]{0.4\textwidth}
  \textbf{Step 2: RAG Evidence Ranking}\\
   \ForEach{generated evidence $d_i \in \mathcal{D}$}{
      \nlset{2}$\text{score}^{(\text{sim})}_i = \cos(\mathbf{e}_i, \mathbf{q})$
      
      \nlset{3}$\text{rank}_{\text{LLM}}(d_i)$ via $\mathcal{M}_\text{large}$
      
      \tcp{Compute combined ranking score}
      \begingroup
      \setlength{\abovedisplayskip}{2pt}
      \setlength{\belowdisplayskip}{2pt}
      \vspace{-3mm}
      \nlset{4}\[
        s_i = \text{score}^{(\text{sim})}_i
              + \text{rank}_{\text{LLM}}(d_i).
      \]
      \endgroup
  }
\nlset{5} Select top-ranked subset $\mathcal{D}_{\text{filtered}} \subset \mathcal{D}$ based on $s_i$.
\end{minipage}
\vspace{0.5em}

\begin{minipage}[t]{0.4\textwidth}
  \textbf{Step 3: Graph RAG Generation}\\
   \ForEach{evidence $d_i \in \mathcal{D}_{\text{filtered}}$}{
      \nlset{6} Prompt $\mathcal{M}_\text{large}$ to extract entity pairs and their relationships
        \begingroup
      \setlength{\abovedisplayskip}{2pt}
      \setlength{\belowdisplayskip}{2pt}
      \[
          \mathcal{R}_i = \{(a, b, r) \mid a, b \in \mathcal{V}, r \in \mathcal{E}\},
      \]
      \endgroup
      where $\mathcal{V}$ is the set of entities, and $\mathcal{E}$ is the set of relationships.

      \nlset{7} Construct $\mathcal{G} = (\mathcal{V}, \mathcal{E})$ by adding nodes $a, b$ and an edge labeled $r$.
  }
 
\end{minipage}
\hfill
\begin{minipage}[t]{0.4\textwidth}
  \textbf{Step 4: Small-Scale LLM Evaluation}\\
   \nlset{8} Select the top $K$ evidences and relationships based on semantic and LLM-based scores.\\ 
  \nlset{9} Prompt $\mathcal{M}_\text{small}$ with:
  \begingroup
      \setlength{\abovedisplayskip}{2pt}
      \setlength{\belowdisplayskip}{2pt}
  \[
    \{ d_i \in \mathcal{D}_{\text{filtered}} \}
    \quad\text{and/or}\quad
    \{(a_j, b_j, r_j)\}_{j=1}^K
  \]
  \endgroup
  to generate the final answer:
  \begingroup
      \setlength{\abovedisplayskip}{2pt}
      \setlength{\belowdisplayskip}{2pt}
  \[
      \hat{y} \leftarrow \mathcal{M}_\text{small}(q,\;\text{context}).
  \]
  \endgroup
  
  \nlset{10} \Return $\hat{y}$ \tcp{Final answer from small LLM.}
\end{minipage}
\end{algorithm*}

\subsection{\name{} Framework Overview}

In this work, we propose \name{} (Distilling RAG for SLMs) as a novel framework to transfer retrieval-augmented generation capabilities from large-scale LLMs to smaller, more efficient models. \name{} mitigates hallucination and enhances answer accuracy by leveraging evidence-based distillation. The overall procedure consists of four sequential steps: {\bf 1)} Evidence generation, {\bf 2)} RAG evidence ranking, {\bf 3)} Graph RAG generation, and {\bf 4)} Small-scale LLM evaluation. A full paradigm of our framework is in Alg.~\ref{alg:mrag}. Each step is described in detail below.

\noindent{\bf Evidence Generation.}
Given an input question $q$, the first stage of \name{} involves eliciting a diverse set of potentially relevant facts from a large-scale language model $\mathcal{M}_{\text{large}}$. Our perspective here is that a well-trained LLM is a stronger and more efficient retriever for SLMs than the traditional ``query encoder + document index based retriever'', especially given the relatively weaker target model. Specifically, we design a prompt (details in our appendix) for $\mathcal{M}_{\text{large}}$ to generate  $N$  distinct textual evidences: $\mathcal{D} = \{ d_1, d_2, \dots, d_{N} \}$.
Each evidence  $d_i$  is intended to encapsulate a factual snippet or a useful piece of information that could help answer $q$. This step not only diversifies the candidate knowledge but also forms the basis for subsequent ranking and structured extraction processes.

\noindent{\bf RAG Evidence Ranking.}
Once evidence set  $\mathcal{D}$  is obtained, each evidence $d_i$ is quantitatively evaluated to determine its relevance to the question $q$. This ranking process involves two key components:\\
\noindent{\bf \em Step-1. Semantic Similarity Score:}
We compute a vector embedding using the sentence-transformers~\cite{reimers2019sentence} 
for both the evidence  $d_i$  and the query  $q$, denoted by $\mathbf{e}_i$ and $\mathbf{q}$ respectively. The semantic similarity score is then calculated using cosine similarity:
\begin{equation}
    \text{score}^{(\text{sim})}_i = \cos(\mathbf{e}_i, \mathbf{q}),
\end{equation}
This score captures the latent semantic alignment between the evidence and the question.

\noindent{\bf \em Step-2. LLM-based Ranking Score:}
In parallel, $\mathcal{M}_{\text{large}}$ is prompted to provide an intrinsic relevance ranking, denoted as $\text{rank}{\text{LLM}}(d_i)$, for each $d_i$. This ranking leverages the vast internal knowledge of $\mathcal{M}_{\text{large}}$ to assess the contextual appropriateness of the evidence.

The combined ranking score $s_i$ for each evidence $d_i$ is computed as an equally weighted sum:
\begin{equation}
    s_i = \text{score}^{(\text{sim})}_i + \text{rank}_{\text{LLM}}(d_i),
\end{equation}
Following the computation of  $s_i$  for all evidences, we discard the lowest-scoring $X$ evidences, where $X$ refers to the $N-K$ portion specified and illustrated in Figure~\ref{fig:methodology}, and retain a filtered subset:
\vspace{-0.1in}
\begin{equation}
    \mathcal{D}_{\text{filtered}} \subset \mathcal{D}.
\end{equation}
This filtering ensures that only the most relevant evidences are carried forward.

\noindent{\bf Graph RAG Generation.} 
In order to further structure the distilled knowledge, the filtered evidences $\mathcal{D}_{\text{filtered}}$ are transformed into a relational graph. For each evidence $d_i \in \mathcal{D}_{\text{filtered}}$, we prompt $\mathcal{M}_{\text{large}}$ to extract structured information in the form of entity relationships. Specifically, for each $d_i$, a set of relationship triples is extracted:
\vspace{-0.05in}
\begin{equation}
    \mathcal{R}_i = \{(a, b, r) \mid a, b \in \mathcal{V},\, r \in \mathcal{E}\},
\vspace{-0.05in}
\end{equation}
where $\mathcal{V}$ represents the set of entities, $\mathcal{E}$ represents the set of relationships.
These triples are then used to construct a graph $\mathcal{G} = (\mathcal{V}, \mathcal{E})$ in which nodes represent entities and edges (labeled by $r$) represent relationships. To focus on the most salient connections, a ranking procedure is applied to the extracted relationships, and the top  $K$  relationships are selected based on a combination of semantic and LLM-based scores (similar to the evidence ranking). This graph-based representation serves as an additional structured context that enriches the evidence pool with inter-entity relationships.

In \name{}, converting evidence into a graph inevitably results in information loss. However, since some evidence pieces are quite long, directly processing them with the SLM would impose a significant computational burden. By utilizing a graph-based representation, we greatly reduce this overhead while preserving essential structured knowledge. To further optimize efficiency, we introduce a {\em simple graph aggregation approach}, as in Figure~\ref{fig:methodology} and Appendix~\ref{simple_graph}, which merges pairs of the same entity into a unified graph representation. This further minimizes computational costs during SLM inference, making the process more efficient without compromising key relational information.

\noindent{\bf SLMs Evaluation.}
In the final step, the distilled and structured evidence is used to inform and boost a small-scale language model $\mathcal{M}_{\text{small}}$ to generate the final answer. The context provided to $\mathcal{M}_{\text{small}}$ can include:

1) The set of filtered evidences: $\{ d_i \in \mathcal{D}_{\text{filtered}}\}$,

2) The top $K$ relationship triples extracted from the graph: $\{ (a_j, b_j, r_j) \}_{j=1}^K$.

These elements are concatenated with the original question $q$ to form a comprehensive prompt. The small-scale model is then queried as follows:
\vspace{-0.05in}
\begin{equation}
    \hat{y} \leftarrow \mathcal{M}_{\text{small}} \bigl(q, \text{context}\bigr).
\vspace{-0.05in}
\end{equation}
where $\hat{y}$ is the final answer. By conditioning on both unstructured evidences and structured relational information, $\mathcal{M}_{\text{small}}$ is better grounded in factual knowledge, thereby mitigating hallucination while maintaining computational efficiency.

\vspace{-0.05in}
\subsection{Mitigating Privacy for Users}
\label{privacy}

Another key advantage of our framework is its potential for privacy protection. Typically, when querying large-scale LLMs, local deployment is not feasible, requiring users to upload their private queries to cloud-based LLMs, raising privacy concerns. Our framework addresses this issue by enabling local SLMs to leverage the knowledge of large models while preserving user privacy.

With our approach, the local model first reformulates the query (much simpler than answering the query directly), stripping any sensitive information before sending it to the cloud-based model. The cloud model then retrieves relevant evidence and knowledge graphs, which are subsequently passed back to the local model for final processing and response generation. This ensures that private data remains protected while still benefiting from the power of large-scale LLMs.

To evaluate the feasibility of this privacy-preserving solution by our \name{} framework, we construct a specialized dataset containing privacy-sensitive information. The dataset includes several key processing steps, with details provided in Appendix~\ref{privacy_algo}.
We test our method on this dataset, as shown in the experimental section, we observe significant improvements in privacy protection while maintaining high accuracy and efficiency.

\subsection{Discussion}

In \name{}, instead of generating answers directly from the teacher model, it solely provides evidence and knowledge graphs for the student model. This strategy offers several key advantages: 1) The teacher LLM is an extremely large, general-purpose model, while the student is domain-specialized, ensuring higher accuracy and efficiency during usage without unnecessary general knowledge. 2) The teacher LLM is usually heavy in size and deployed on the cloud, and the student is on local devices, it is simple to develop methods for preserving privacy using our proposed framework by transmitting only de-identified queries and structured knowledge instead of full responses, as we have introduced in Section~\ref{privacy}.

\begin{table*}[h]
\centering
\resizebox{0.8\linewidth}{!}{
    \begin{tabular}{lcccc}
    \toprule
    \multirow{2}{*}{\textbf{Framework}}  & \multirow{2}{*}{\textbf{Backbone LLM}} &
    \textbf{MedMCQA} &
    \textbf{MMLU} &
    \textbf{ARC-C}\\
    
    &  &Acc & Acc & Acc   \\
    \midrule
    Self-RAG~\citep{selfrag} & SelfRAG-LLaMA-2-7B & -- & -- & \ 67.3  \\
    CRAG~\citep{crag} & SelfRAG-LLaMA-2-7B & -- & -- & \ 68.6 \\
    \name{} (Ours) & LLaMA-2-7B & 72.4 & 71.2 & \quad \quad 86.2\textcolor{table1blue}{$_{\bf\uparrow17.6}$}  \\  \hdashline
    SimRAG~\citep{simrag} & Llama-3.1-8B-Instruct & --  & 67.5 & \ 81.4 \\ 
    \name{} (Ours) &  Llama-3.1-8B-Instruct  & 74.2 & \quad \ 75.7\textcolor{table1blue}{$_{\bf\uparrow8.2}$} & \quad \quad 93.1\textcolor{table1blue}{$_{\bf\uparrow11.7}$} \\
    \midrule
    & GLM-edge-1.5B-chat &52.1 & 47.0 & \ 62.3 \\ 
    & Qwen2.5-3B-Instruct &58.3 & 70.9 & \ 67.7 \\
    MiniRAG~\citep{minirag}& Llama-3.2-3B-Instruct & 52.7 & 69.1 & \ 65.3 \\
    & Gemma-2-2B-it &48.5 & 57.3 & \ 68.6 \\
    & Phi-3.5-mini-instruct &61.1 & 72.7 & \ 82.7 \\
    \midrule
    \multirow{5}{*}{\name{} (Ours)}     & GLM-edge-1.5B-chat &	\quad 56.9\textcolor{table1blue}{$_{\bf\uparrow4.8}$} 	& \quad \ \ \ 69.0\textcolor{table1blue}{$_{\bf\uparrow22.0}$} 	& \quad \ \ \ 90.0\textcolor{table1blue}{$_{\bf\uparrow27.7}$}     \\
    & Qwen2.5-3B-Instruct &\quad \ \ 72.8\textcolor{table1blue}{$_{\bf\uparrow14.5}$}    &\quad\ \  73.8\textcolor{table1blue}{$_{\bf\uparrow2.9}$}  &\quad\quad 93.0\textcolor{table1blue}{$_{\bf\uparrow25.3}$}     \\
     & Llama-3.2-3B-Instruct &\quad\ \ 73.6\textcolor{table1blue}{$_{\bf\uparrow20.9}$}    &\quad\ \  74.4\textcolor{table1blue}{$_{\bf\uparrow5.3}$}    & \quad\quad 93.0\textcolor{table1blue}{$_{\bf\uparrow27.7}$}     \\
    & Gemma-2-2B-it &\quad\ \ 72.4\textcolor{table1blue}{$_{\bf\uparrow23.9}$}   &\quad \quad 71.2\textcolor{table1blue}{$_{\bf\uparrow13.9}$}   & \quad\quad 91.5\textcolor{table1blue}{$_{\bf\uparrow22.9}$}    \\
    & Phi-3.5-mini-instruct &\quad\ \ 74.4\textcolor{table1blue}{$_{\bf\uparrow13.3}$}   & \quad\ \ 77.8\textcolor{table1blue}{$_{\bf\uparrow5.1}$}    & \quad\quad94.1\textcolor{table1blue}{$_{\bf\uparrow11.4}$}     \\
    \bottomrule
    \end{tabular}
}
\vspace{-0.1in}
\caption{\textbf{Comparison with other state-of-the-art RAG frameworks.} We compare \name{} (evidence-based) with prior approaches across multiple benchmarks and backbone LLMs.
In the table, the ``\textcolor{table1blue}{$\uparrow$}''  indicates improvements over other methods under the same backbone and inference configuration.}  
\label{compare_sota}
\vspace{-0.1in}
\end{table*}

\section{Experiments}

\noindent{\bf Datasets.} 
The following benchmarks are used in our work:
\textbf{ARC-Challenge} \cite{clark2018thinksolvedquestionanswering}, 
\textbf{MedMCQA} \cite{pal2022medmcqalargescalemultisubject},
\textbf{GPQA} \citep{rein2023gpqagraduatelevelgoogleproofqa},
{\bf MMLU} \citep{hendrycks2021measuringmassivemultitasklanguage},
{\bf Open-LLM-Leaderboard} \citep{myrzakhan2024openllmleaderboardmultichoiceopenstylequestions},
{\bf AVERITEC} \citep{schlichtkrull2023averitecdatasetrealworldclaim}.
More details of these datasets are provided in Appendix~\ref{datasets}.

\vspace{-0.05in}
\subsection{\bf Models and Experimental Settings}

For our experiment we use a set of large-scale teacher models and small student models. The teacher models include GPT-4o \cite{hurst2024gpt}, DeepSeek-V3 \cite{liu2024deepseek}, Gemini Flash 1.5 \cite{team2024gemini}, Claude 3.5 Sonnet \cite{anthropic2024claude35}, and LLaMA-3.3-70B \cite{dubey2024llama}. For student models, we use Gemma-2-2B-it \cite{team2024gemma}, Phi-3.5-mini-instruct \cite{abdin2024phi}, Qwen2.5-3B-Instruct \cite{yang2024qwen2}, LLaMA-3.2-3B-Instruct \cite{dubey2024llama}, Qwen2.5-7B-Instruct \cite{yang2024qwen2}, LLaMA-3.1-8B-Instruct \cite{dubey2024llama}, and  Gemma-2-9B-it \cite{team2024gemma} covering a range of 2B to 9B parameters. We evaluate the performance of the student models in a zero-shot setting using the lm-evaluation-harness framework \cite{eval-harness} on a 4$\times$RTX 4090 GPUs setup.

\vspace{-0.02in}
\subsection{Comparison with State-of-the-Arts}

Table~\ref{compare_sota} compares \name{} against existing RAG frameworks on MedMCQA, MMLU, and ARC-C. Overall, \name{} consistently outperforms previous state-of-the-art methods and effectively boosts small language models (SLMs). For instance, on ARC-C, Self-RAG~\citep{selfrag} and CRAG~\citep{crag} achieve 67.3\% and 68.6\%, respectively, while \name{} obtains up to 94.1\%, exceeding them by +26.8\% and +25.5\%. 
SimRAG~\citep{simrag}, based on the Llama-3.1-8B-Instruct backbone, achieves 67.5\% and 81.4\% on MMLU and ARC-C, respectively. By contrast, \name{} attains 75.7\% and 93.1\% with the same backbone, surpassing SimRAG by +8.2\%, and +11.7\%. 
Compared to MiniRAG~\citep{minirag}, \name{} achieves notable gains of at most +23.9\% on MedMCQA, +13.9\% on MMLU, and +11.4\% on ARC-C with the same SLMs backbones. These results confirm that \name{} delivers superior retrieval-augmented performance while substantially reducing hallucination and computational overhead.

\subsection{Ablation Studies} 

\begin{figure*}[t]
\centering
\includegraphics[width=1.\linewidth]{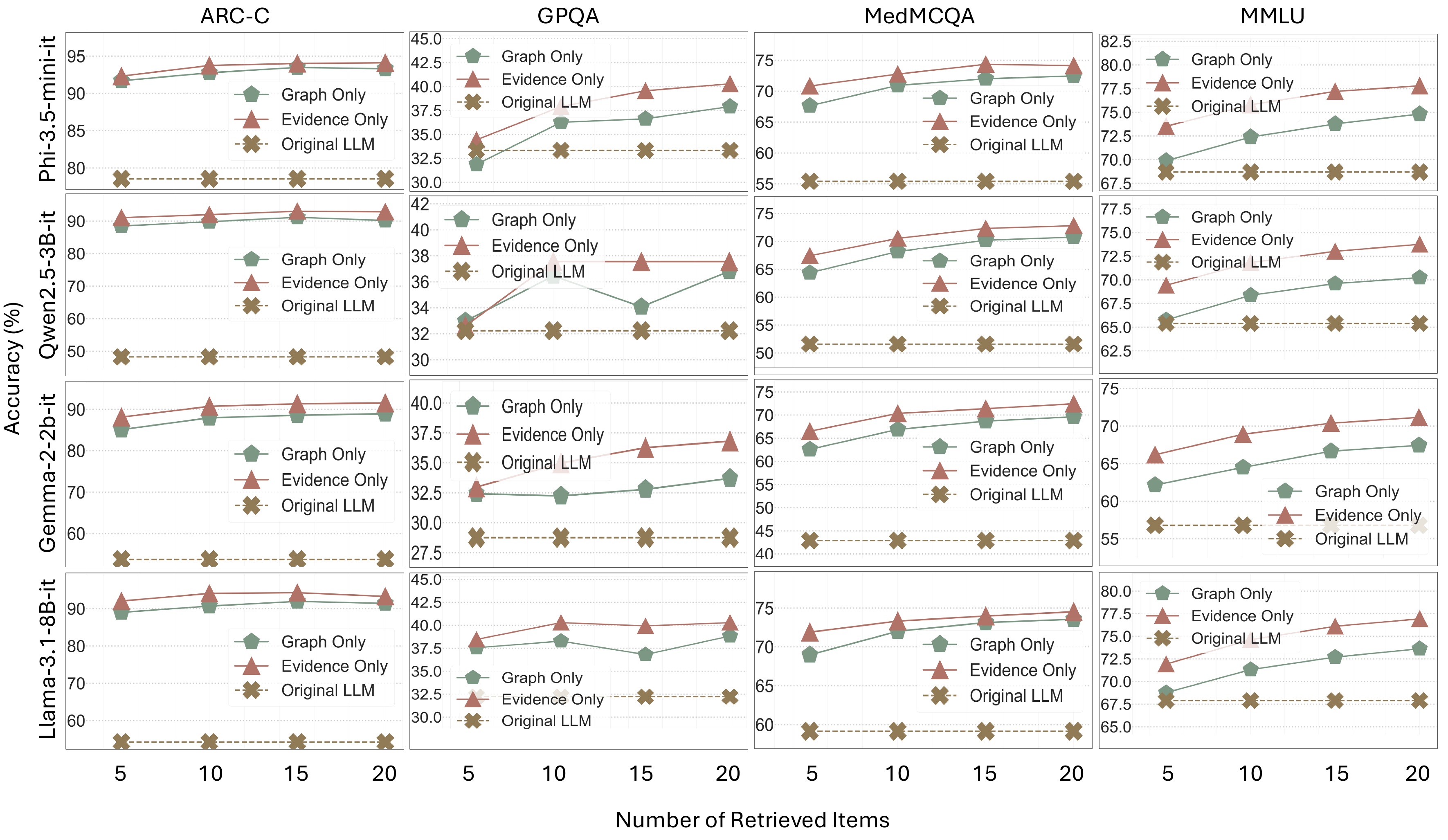}  
\vspace{-0.3in}
\caption{\textbf{Effect of retrieved graph-based and evidence-based RAG on multiple-choice QA tasks.} We evaluate different retrieval strategies: {\em Graph Only}, {\em Evidence Only}, and {\em the Original LLM} across four benchmarks (ARC-C, GPQA, MedMCQA, and MMLU) using various backbone models. The x-axis represents the number of retrieved items, while the y-axis denotes accuracy (\%).}
\label{methods_ablation}
\vspace{-0.15in}
\end{figure*}

\noindent{\bf Number of evidence/graph relations $K$.} We analyze the impact of varying the number of generated/retrieved evidence on student SLM's performance, as in Table~\ref{tab:model-performance-gpt-arc} and Figure~\ref{methods_ablation}. Our experiments show that $\sim$15 evidence pieces generally provide the optimal cost-accuracy trade-off results, using fewer leads to insufficient knowledge, while using more introduces redundancy and slightly degrades performance due to increased noise. Since the graph representation is constructed from raw evidence, it naturally loses information but remains more concise and computationally efficient. Combining both evidence and graph might be beneficial, our results show that this scheme is redundant, yielding similar accuracy to using evidence alone, while incurring extra inference overhead.

\noindent{\bf Effects of different teacher LLMs.}  We investigate the effect of different teacher LLMs on student performance as in Table~\ref{tab:fact_verification_multimodel__different_teachers}. Surprisingly, a more powerful teacher does not always lead to better student accuracy. Our experiments show that GPT-4o produces the best distillation results, outperforming all other models. The ranking of teacher models for student SLM is as follows: GPT-4o $>$ Claude 3.5 Sonnet $>$ DeepSeek V3 $>$ Llama 3.3 70B $>$ Gemini 1.5 Flash. These results indicate that the quality and consistency of generated evidence matter more than just using a more capable LLM. Certain models, such as Claude and DeepSeek V3, perform competitively, but their evidence generation may not be as structured or factually aligned as GPT-4o.

\begin{table*}[t]
    \centering
    \resizebox{\textwidth}{!}{ 
        \begin{tabular}{lc|cccc|cccc|cccc}
            \toprule
            \multirow{2}{*}{\textbf{Target SLM}} & \multirow{2}{*}{\textbf{Original}} 
            & \multicolumn{4}{c|}{\textbf{Graph Only}} 
            & \multicolumn{4}{c|}{\textbf{Evidence Only}} 
            & \multicolumn{4}{c}{\textbf{Graph and Evidence Combined}} \\
            \cmidrule(lr){3-6} \cmidrule(lr){7-10} \cmidrule(lr){11-14}
            &  & 5 & 10 & 15 & 20
            & 5 & 10 & 15 & 20
            & 5 & 10 & 15 & 20 \\
            \midrule
            Phi-3.5-mini-instruct & 78.55 & 91.69 & 92.76 & 93.48 & 93.30 & 92.31 & 93.74 & 94.01 & 94.10 & 92.40 & 93.12 & 93.74 & 93.74 \\
            Qwen2.5-3B-Instruct   & 48.26 & 88.47 & 89.81 & 91.15 & 90.17 & 91.06 & 91.96 & 93.03 & 92.85 & 89.81 & 91.06 & 92.49 & 92.23 \\
            Llama-3.2-3B-Instruct & 42.45 & 86.77 & 88.47 & 90.44 & 90.44 & 89.10  & 91.78 & 92.31 & 93.03 & 88.56 & 91.51 & 92.67 & 92.40  \\
            Llama-3.1-8B-Instruct & 54.24 & 89.01 & 90.71 & 91.96 & 91.42 & 92.05 & 94.10  & 94.28 & 93.30  & 91.69 & 93.03 & 94.28 & 93.57 \\
            Qwen2.5-7B-Instruct   & 55.23 & 90.97 & 91.78 & 92.49 & 92.58 & 91.87 & 93.12 & 93.39 & 93.83 & 91.51 & 93.03 & 93.48 & 93.74 \\
            gemma-2-9b-it        & 63.27 & 92.58 & 93.30  & 93.74 & 94.10  & 93.74 & 94.28 & 94.73 & 94.46 & 93.30  & 94.19 & 94.28 & 94.37 \\
            gemma-2-2b-it        & 53.71 & 85.08 & 87.94 & 88.56 & 88.92 & 88.11 & 90.71 & 91.33 & 91.51 & 87.67 & 90.71 & 91.33 & 91.33 \\
            \bottomrule
        \end{tabular}
    }
    \vspace{-0.1in}
    \caption{\textbf{Comparison of results on the ARC-Challenge using GPT-4o as the teacher model}. The {\em Original} represents the baseline performance without RAG. {\em Evidence Only} uses ranked context textual evidences, {\em Graph Only} utilizes structured relationships, and {\em Graph and Evidence Combined} integrates both sources. Results are reported for different retrieval sizes (5, 10, 15, 20).}
    \label{tab:model-performance-gpt-arc}
    \vspace{0.1in}

        \centering
    \resizebox{\textwidth}{!}{
        \begin{tabular}{lc|ccc|ccc|ccc|ccc|ccc}
            \hline
            \multirow{2}{*}{\textbf{Target SLM}} & \multirow{2}{*}{\textbf{Original}} & \multicolumn{3}{c}{\textbf{GPT-4o}} & \multicolumn{3}{c}{\textbf{Llama 3.3 70b}} & \multicolumn{3}{c}{\textbf{Claude 3.5 Sonnet}} & \multicolumn{3}{c}{\textbf{Gemini 1.5 Flash}} & \multicolumn{3}{c}{\textbf{DeepSeek V3}} \\
             &  & \textbf{Graph} & \textbf{Evide.} & \textbf{Comb.} & \textbf{Graph} & \textbf{Evide.} & \textbf{Comb.} & \textbf{Graph} & \textbf{Evide.} & \textbf{Comb.} & \textbf{Graph} & \textbf{Evide.} & \textbf{Comb.} & \textbf{Graph} & \textbf{Evide.} & \textbf{Comb.} \\
            \hline
            Phi-3.5-mini-instruct & 55.41 & 72.01 & 74.35 & 74.06 & 65.43 & 68.80 & 68.42 & 65.14 & 69.78 & 70.28 & 58.91 & 61.18 & 60.03 & 66.12 & 68.95 & 68.49 \\
            Qwen2.5-3B-Instruct & 51.59 & 70.21 & 72.32 & 72.03 & 61.06 & 66.48 & 65.34 & 60.32 & 65.50 & 67.68 & 55.08 & 57.66 & 57.04 & 63.18 & 66.27 & 66.08 \\
            Llama-3.2-3B-Instruct & 50.90 & 71.65 & 73.56 & 73.15 & 64.26 & 67.99 & 67.34 & 64.69 & 68.56 & 69.21 & 59.62 & 61.58 & 60.60 & 66.56 & 69.16 & 68.80 \\
            Llama-3.1-8B-Instruct & 59.14 & 73.13 & 73.97 & 73.54 & 67.34 & 69.52 & 69.02 & 68.95 & 71.74 & 71.69 & 62.99 & 63.78 & 63.09 & 68.35 & 70.40 & 69.4 \\
            Qwen2.5-7B-Instruct & 56.08 & 72.51 & 73.66 & 73.58 & 64.04 & 68.13 & 67.54 & 65.67 & 70.07 & 70.14 & 59.84 & 60.77 & 60.58 & 66.56 & 68.76 & 68.54 \\
            gemma-2-9b-it & 56.83 & 72.79 & 74.13 & 73.73 & 66.39 & 69.33 & 69.23 & 69.38 & 71.34 & 71.48 & 61.34 & 61.65 & 61.32 & 68.06 & 69.59 & 69.42 \\
            gemma-2-2b-it & 42.91 & 68.68 & 71.38 & 71.72 & 60.79 & 65.54 & 65.19 & 57.76 & 63.93 & 66.65 & 54.24 & 57.21 & 56.32 & 62.20 & 66.24 & 65.46 \\
            \hline
        \end{tabular}
    }
    \vspace{-0.1in}
    \caption{Ablation comparison across various large-scale teacher models on MedMCQA.}
    \label{tab:fact_verification_multimodel__different_teachers}
    \vspace{0.1in}
    
    \centering
    \resizebox{\textwidth}{!}{
        \begin{tabular}{lc|cccc|cccc|cccc}
            \toprule
            \multirow{2}{*}{\textbf{Target SLM}} & \multirow{2}{*}{\textbf{Original}} 
            & \multicolumn{4}{c|}{\textbf{Graph Only}} 
            & \multicolumn{4}{c|}{\textbf{Evidence Only}} 
            & \multicolumn{4}{c}{\textbf{Graph and Evidence Combined}} \\
            \cmidrule(lr){3-6} \cmidrule(lr){7-10} \cmidrule(lr){11-14}
            &  & 5 & 10 & 15 & 20
            & 5 & 10 & 15 & 20
            & 5 & 10 & 15 & 20 \\
            \midrule
            Phi-3.5-mini-instruct & 66.22 & 69.24 & 70.35 & 71.46 & 71.97 & 72.11 & 74.48 & 75.13 & 75.58 & 72.22 & 74.05 & 74.46 & 74.48 \\
            Qwen2.5-3B-Instruct   & 62.01 & 64.89 & 66.73 & 67.29 & 68.17 & 68.81 & 70.78 & 71.50 & 71.95 & 67.97 & 70.41 & 70.58 & 71.58 \\
            Llama-3.2-3B-Instruct & 57.27 & 63.20 & 64.93 & 65.38 & 66.76 & 67.54 & 69.98 & 70.82 & 71.36 & 67.49 & 69.51 & 70.10 & 71.13 \\
            Llama-3.1-8B-Instruct & 65.01 & 67.78 & 68.95 & 69.92 & 70.70 & 70.94 & 73.29 & 73.96 & 74.15 & 71.15 & 72.83 & 73.31 & 73.39 \\
            Qwen2.5-7B-Instruct   & 69.71 & 70.49 & 72.55 & 73.27 & 73.65 & 73.65 & 75.59 & 76.30 & 76.34 & 73.72 & 75.75 & 75.95 & 76.16 \\
            gemma-2-9b-it         & 69.73 & 71.25 & 73.35 & 73.82 & 74.68 & 73.31 & 75.30 & 76.32 & 76.49 & 73.74 & 75.91 & 76.16 & 76.49 \\
            gemma-2-2b-it         & 55.59 & 59.61 & 61.66 & 62.50 & 63.45 & 64.25 & 66.16 & 66.94 & 67.58 & 63.63 & 66.12 & 66.59 & 66.88 \\
            \bottomrule
        \end{tabular}
    }
    \vspace{-0.1in}
    \caption{Performance of using GPT-4o as the teacher on privacy protection task and benchmark.} 
    \label{tab:model-performance-privacy}
    \vspace{-0.1in}
\end{table*}

\noindent{\bf Computation comparison.} To compare computation efficiency, we evaluate the average length of generated evidence versus knowledge graphs in Table~\ref{tab:computation_results}. As expected, graph-based RAG significantly reduces computational costs, as the structured representation is much shorter than raw evidence while still preserving core relational information. Specifically, graph representations require significantly fewer tokens during inference, making them ideal for low-resource or real-time retrieval scenarios.

\begin{table}[h]
    \centering
    \vspace{-0.06in}
    \resizebox{0.4\textwidth}{!}{ 
    \begin{tabular}{l|c|c}
        \toprule
        Category & Total Number & Average Length \\
        \hline
        Evidence & 29,698,547 & 26.30 \\
        Graph & 24,324,636 & \quad \quad \ \  21.55\textcolor{table1blue}{$_{\bf\downarrow18.1\%}$}    \\
        \bottomrule
    \end{tabular}
    }
        \vspace{-0.13in}
    \caption{Token statistics per evidence and graph.}
    \label{tab:computation_results}
    \vspace{-0.15in}
\end{table}

\subsection{Multi-choice QA}

To assess the impact of \name{} on retrieval-augmented multiple-choice question answering (MCQA), we conduct extensive experiments across four benchmark datasets: ARC-C, MedMCQA, GPQA, and MMLU in Table~\ref{tab:model-performance-gpt-arc} and Tables~\ref{tab:model-performance-gpt-med},~\ref{tab:model-performance-gpt-gpqa}, and~\ref{tab:model-performance-gpt-mmlu} in Appendix, testing various students and with GPT-4o as the teacher model. The results demonstrate significant improvement across various student and teacher model architectures.

Among student models, Gemma-2-9b-it consistently achieves the strongest performance across benchmarks when paired with GPT-4o as the teacher, reaching 94.73\% on ARC-C, 77.80\% on MMLU, 74.42\% on MedMCQA, and 40.11\% on GPQA with evidence-only distillation. This represents substantial improvements over baseline performance: 53.71\%, 71.80\%, 56.83\%, and 34.80\%, respectively. The Phi-3.5-mini-instruct model, despite its smaller size, shows surprisingly competitive results, particularly on ARC-C (94.10\%) and MMLU (77.80\%). In contrast, smaller variants like Gemma-2-2b-it and Llama-3.2-3B-Instruct consistently perform 3-5\% lower than their larger counterparts, although still showing substantial improvements over their baselines. In particular, on MMLU, Gemma-2b-it improves from 56.80\% to 71.16\%, demonstrating effective knowledge transfer even to resource-constrained architectures.

\subsection{Open-ended QA}
To assess the effectiveness of \name{} on open-style questions, where it requires models to generate unconstrained, contextually appropriate responses rather than selecting from predefined choices. We used the Open-LLM Leaderboard \cite{myrzakhan2024openllmleaderboardmultichoiceopenstylequestions} for evaluation. Given the large scale of the Open-LLM Leaderboard, we considered computational cost and opted for GPT-4o-mini \cite{openai2024gpt4omini} as the teacher model, balancing efficiency with effective knowledge transfer.

\begin{table}[t]
    \centering
    \vspace{0.09in}
    \resizebox{0.48\textwidth}{!}{    
    \begin{tabular}{lcccc}
        \toprule
        \textbf{Open-LLM Leaderboard} & \textbf{Original} &  \bf {\name{}$_{G}$} &  \bf {\name{}$_{E}$} &  \bf {\name{}$_{C}$} \\
        \midrule
        Phi-3.5-mini-instruct  & 35.26  & 38.22\textcolor{table1blue}{$_{\bf\uparrow2.96}$}     & 40.13\textcolor{table1blue}{$_{\bf\uparrow 4.87}$}    & 39.84\textcolor{table1blue}{$_{\bf\uparrow 4.58}$}    \\
        Qwen2.5-3B-Instruct    & 32.59  & 36.01\textcolor{table1blue}{$_{\bf\uparrow 3.42}$}    & 37.23\textcolor{table1blue}{$_{\bf\uparrow 4.64}$}    & 37.50\textcolor{table1blue}{$_{\bf\uparrow 4.91}$}    \\
        Llama-3.2-3B-Instruct  & 29.85  & 32.79\textcolor{table1blue}{$_{\bf\uparrow 2.94}$}    & 34.12\textcolor{table1blue}{$_{\bf\uparrow 4.27}$}    & 33.65\textcolor{table1blue}{$_{\bf\uparrow 3.80}$}     \\
        Llama-3.1-8B-Instruct  & 45.07  & 49.13\textcolor{table1blue}{$_{\bf\uparrow 4.06}$}    & 50.26\textcolor{table1blue}{$_{\bf\uparrow 5.19}$}    & 51.73\textcolor{table1blue}{$_{\bf\uparrow 6.66}$}    \\
        Qwen2.5-7B-Instruct    & 44.67  & 48.73\textcolor{table1blue}{$_{\bf\uparrow 4.06}$}    & 52.36\textcolor{table1blue}{$_{\bf\uparrow 7.69}$}    & 52.66\textcolor{table1blue}{$_{\bf\uparrow 7.99}$}    \\
        gemma-2-9b-it         & 46.44  & 49.12\textcolor{table1blue}{$_{\bf\uparrow 2.68}$}     & 53.45\textcolor{table1blue}{$_{\bf\uparrow 7.01}$}    & 53.19\textcolor{table1blue}{$_{\bf\uparrow 6.75}$}    \\
        gemma-2-2b-it         & 32.54  & 35.98\textcolor{table1blue}{$_{\bf\uparrow 3.44}$}     & 37.31\textcolor{table1blue}{$_{\bf\uparrow 4.77}$}    & 37.26\textcolor{table1blue}{$_{\bf\uparrow 4.72}$}     \\
        \bottomrule
    \end{tabular}
    }
    \vspace{-0.1in}
    \caption{Accuracy on Open-LLM leaderboard. {\name{}$_{G}$}, {\name{}$_{E}$}, and {\name{}$_{C}$} represent graph-based, evidence-based and combined configurations, respectively.}
    \label{tab:llm_results_osq}
    \vspace{-0.15in}
\end{table}

As shown in Table \ref{tab:llm_results_osq}, both graph-based and evidence-based distillation lead to significant improvements over the original model performances, where the original refers to student models without any distillation. Evidence-based setting provides the highest accuracy gains, with Gemma-2-9b-it improving from 46.44\% to 53.45\%, and Qwen2.5-7B-Instruct from 44.67\% to 52.36\%. These results highlight the importance of structured knowledge and retrieved evidence in enhancing open-style response generation. Graph-only distillation, while slightly less effective, still provides meaningful improvements, where Qwen2.5-3B-Instruct increases from 32.59\% to 36.01\%. These results emphasize the importance of utilizing both structured and retrieved information to improve open-style response generation while demonstrating that smaller, cost-effective teacher models like GPT-4o-mini can still facilitate meaningful knowledge transfer.

\subsection{Fact Verification}

\begin{table}[h]
    \centering
    \resizebox{0.48\textwidth}{!}{
        \begin{tabular}{lcccc}
            \hline
            \textbf{Model} & \textbf{Original} & \bf {\name{}$_{G}$} & \bf {\name{}$_{E}$} & \bf {\name{}$_{C}$} \\
            \hline
            BLOOM-7b & 26 & \ \ 30.11\textcolor{table1blue}{$_{\bf\uparrow 4.11}$}   & \ \ 32.43\textcolor{table1blue}{$_{\bf\uparrow 6.43}$}   & \ \ 32.29\textcolor{table1blue}{$_{\bf\uparrow 6.29}$}   \\
            GPT-3.5-Turbo & 29 & \ \ \ 40.98\textcolor{table1blue}{$_{\bf\uparrow 11.98}$}    & \ \ \ 49.10\textcolor{table1blue}{$_{\bf\uparrow 20.10}$}   & \ \ \ 45.63\textcolor{table1blue}{$_{\bf\uparrow 16.63}$}    \\
            \hline
        \end{tabular}
    }
    \vspace{-0.1in}
    \caption{Performance on AVERITEC.} 
    \label{tab:fact_verification}
    \vspace{-0.1in}
\end{table}

 Following ~\citet{schlichtkrull2023averitecdatasetrealworldclaim}, BLOOM-7b and GPT-3.5-Turbo are used as students for fact verification benchmarking on AVERITEC dataset as in Table~\ref{tab:fact_verification}. For both models, evidence distillation yielded the strongest performance, with 32.43\% for BLOOM-7b and 49.10\% for GPT-3.5-Turbo. Evidence and graph combined distillation provided the second highest in both cases, with 32.29\% for BLOOM-7b and 45.63\% for GPT-3.5-Turbo.

\subsection{Privacy Protection Evaluation}
Our privacy-preserving framework effectively minimizes the risk of sensitive data exposure. We analyze the reduction in personally identifiable information (PII) before and after applying our SLM-based privacy filtering. Out of 15,090 injected PIIs, only 649 remain post-processing, resulting in an overall reduction of 95.7\%.

To understand the impact of privacy filtering on model accuracy, we evaluate performance on our MMLU-augmented dataset (see Appendix \ref{privacy_algo}). As shown in Table \ref{tab:model-performance-privacy}, our framework maintains strong performance across various student models, despite rigorous privacy filtering. The graph and evidence-based combined approach achieves the best results, with Gemma-2-9b-it increasing from 69.73\% to 76.49\% and Qwen2.5-7B-Instruct improving from 69.71\% to 76.16\%. Even smaller models like Gemma-2-2b-it show notable gains, rising by 11.29\% from the baseline, demonstrating that privacy filtering does not significantly compromise the performance of \name{}. These findings confirm that our framework effectively mitigates privacy risks while preserving knowledge retrieval, ensuring high-quality LLM responses.

\section{Conclusion}

We presented \name{}, a novel approach that distills RAG knowledge into SLMs using evidence- and graph-guided distillation. By structuring knowledge extraction with ranked evidence and knowledge graphs, \name{} mitigates hallucinations while significantly reducing computational demands. Experimental results show that \name{} outperforms existing SoTAs like MiniRAG under similar constraints, preserving RAG's benefits while enhancing efficiency. Our work offers a scalable, resource-efficient solution for deploying high-quality retrieval-augmented generation in small models, balancing factual consistency and computational efficiency in knowledge-intensive tasks.

\section*{Limitations}

Despite its strong performance, \name{} has a few limitations that warrant further investigation: 1) {\em Knowledge retention trade-offs.} Our method successfully distills factual knowledge into smaller models, but some nuanced or implicit knowledge present in the teacher LLMs may be missing. This is especially relevant in creative, open-ended, or subjective tasks where explicit factual grounding is less defined. 2) {\em Computational overhead during distillation.} Although \name{} enables more efficient inference in SLMs, the distillation process itself requires significant computation, particularly when generating evidence rankings and graph-based knowledge representation. Future work could explore optimizing this process to further reduce evidence generation costs. 3) In \name{}, when generating evidence, we aim to prevent data/answer leakage by instructing the model explicitly in the prompt with ``do not give the answer away directly''. However, despite this precaution, there is still a potential risk of unintended leakage. This could raise concerns regarding the integrity of the distillation process. To mitigate this, we ensure that the generated evidence remains neutral, contextually relevant, and free from direct answer hints while still being informative for the target student model.

\section*{Ethics Statement}

While \name{} minimizes hallucinations, its outputs must still be subject to critical evaluation in high-stakes applications such as legal, medical, or scientific domains. Human oversight remains crucial in ensuring that AI-generated content aligns with ethical and professional standards.
Also, \name{} reduces hallucinations by aligning outputs with structured knowledge, but it remains susceptible to biases present in the teacher LLMs, knowledge graphs, and evidence. Biases inherent in training data may still propagate into distilled SLMs, necessitating continuous evaluation and mitigation strategies.

\bibliography{custom}
\bibliographystyle{acl_natbib}

\clearpage

\appendix

\section*{\Large{Appendix}}

\section{Datasets} \label{datasets}

The following benchmarks are used in our work, more details of them are as follows:

\noindent\textbf{ARC-Challenge}: The AI2 Reasoning Challenge (ARC) is comprised of questions pertaining to natural science \cite{clark2018thinksolvedquestionanswering}. The benchmark is split into the Easy Set and the Challenge Set, we selected ARC-Challenge due to the fact that it consists of questions for which retrieval and co-occurrence methods both fail.

\noindent\textbf{MedMCQA}: It contains over 194k medical school entrance exam questions spanning over 21 medical subjects \cite{pal2022medmcqalargescalemultisubject}. This is a high-quality benchmark with real medical exam questions.

\noindent\textbf{GPQA}: This is the Google-Proof QA dataset, a collection of 448 multiple-choice questions written by experts in the natural sciences. What makes this dataset unique is that PhDs and PhD candidates in the corresponding domains have only reached 65\% accuracy in the questions even with web access, hence making the questions "Google-proof" \citep{rein2023gpqagraduatelevelgoogleproofqa}. 

\noindent{\bf MMLU}: The {\em Massive Multitask Language Understanding} benchmark is an evaluation dataset consisting of multiple-choice tasks covering a broad range of 57 subjects \citep{hendrycks2021measuringmassivemultitasklanguage}. MMLU is used for its comprehensiveness and high popularity level.

\noindent{\bf Open-LLM-Leaderboard}: To circumvent issues associated with multiple choice questions, such as preference towards certain options, we include an open-style benchmark in our evaluation that avoids the selection bias and random guessing problems \citep{myrzakhan2024openllmleaderboardmultichoiceopenstylequestions}. 

\noindent{\bf AVERITEC}: It measures LLM fact verification abilities by providing claims that LLMs either refute, support, claim not enough evidence, or claim conflicting evidence \citep{schlichtkrull2023averitecdatasetrealworldclaim}.

\begin{figure*}[t]
\centering
\includegraphics[width=1.\linewidth]{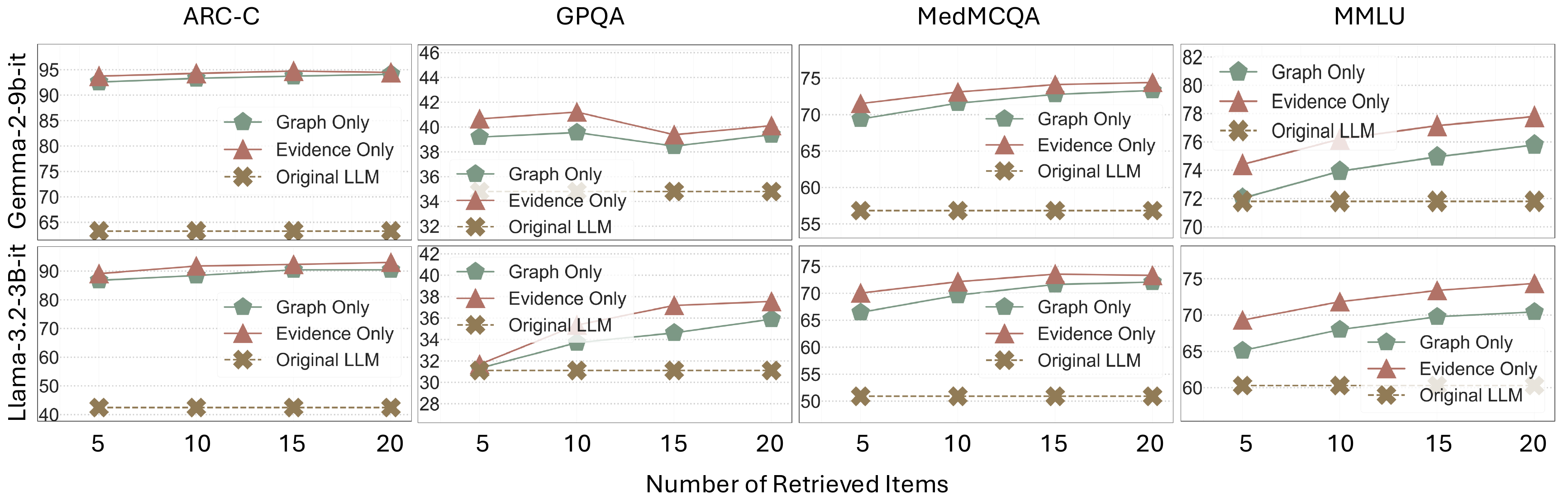}  
\vspace{-0.25in}
\caption{\textbf{More results on Retrieval Strategies.} We evaluate different retrieval strategies: {\em Graph Only}, {\em Evidence Only}, and {\em the Original LLM} across four benchmarks (ARC-C, GPQA, MedMCQA, and MMLU) extended on Llama3.2-3B-it and Gemma-2-9B-it benchmarks. In this figure, the x-axis represents the number of retrieved items, while the y-axis denotes accuracy (\%).}
\label{methods_ablation_appendix}
\end{figure*}

\begin{table*}[h]
    \centering
    \resizebox{\textwidth}{!}{ 
        \begin{tabular}{lc|cccc|cccc|cccc}
            \toprule
            \multirow{2}{*}{\textbf{Target SLM}} & \multirow{2}{*}{\textbf{Original}} 
            & \multicolumn{4}{c|}{\textbf{Graph Only}} 
            & \multicolumn{4}{c|}{\textbf{Evidence Only}} 
            & \multicolumn{4}{c}{\textbf{Graph and Evidence Combined}} \\
            \cmidrule(lr){3-6} \cmidrule(lr){7-10} \cmidrule(lr){11-14}
            &  & 5 & 10 & 15 & 20
            & 5 & 10 & 15 & 20
            & 5 & 10 & 15 & 20 \\
            \midrule
            Phi-3.5-mini-instruct & 55.41 & 67.65 & 70.93 & 72.01 & 72.46 & 70.83 & 72.77 & 74.35 & 74.13 & 69.73 & 72.36 & 74.06 & 74.06 \\
            Qwen2.5-3B-Instruct   & 51.59 & 64.38 & 68.18 & 70.21 & 70.76 & 67.42 & 70.55 & 72.32 & 72.82 & 66.75 & 69.95 & 72.03 & 72.13 \\
            Llama-3.2-3B-Instruct & 50.90 & 66.46 & 69.66 & 71.65 & 72.08 & 70.05 & 72.15 & 73.56 & 73.32 & 69.47 & 71.62 & 73.15 & 73.27 \\
            Llama-3.1-8B-Instruct & 59.14 & 68.97 & 72.01 & 73.13 & 73.56 & 71.89 & 73.32 & 73.97 & 74.52 & 71.58 & 72.96 & 73.54 & 74.18 \\
            Qwen2.5-7B-Instruct   & 56.08 & 67.44 & 70.43 & 72.51 & 72.53 & 70.48 & 72.56 & 73.66 & 74.01 & 69.54 & 72.24 & 73.58 & 73.68 \\
            gemma-2-9b-it         & 56.83 & 69.40 & 71.58 & 72.79 & 73.32 & 71.53 & 73.11 & 74.13 & 74.42 & 71.34 & 72.89 & 73.73 & 73.97 \\
            gemma-2-2b-it         & 42.91 & 62.59 & 66.91 & 68.68 & 69.64 & 66.51 & 70.36 & 71.38 & 72.44 & 65.62 & 69.35 & 71.72 & 72.03 \\
            \bottomrule
        \end{tabular}
    }
    \vspace{-0.1in}
    \caption{Comparison of results on the MedMCQA using GPT-4o as the teacher model.}
    \label{tab:model-performance-gpt-med}
    \vspace{-0.06in}
\end{table*}

\begin{table}[t]
    \centering
    \vspace{0.05in}
    \resizebox{0.48\textwidth}{!}{
        \begin{tabular}{lcccc}
            \hline
            \textbf{Model} & \textbf{ARC-C} & \textbf{MedMCQA} & \textbf{GPQA} & \textbf{MMLU} \\
            \hline
            GPT-4o & 96.7 & 77.0 & 53.6 & 88.7 \\
            Claude 3.5 Sonnet & 96.4 & 76.1 & 59.4 & 88.7 \\
            DeepSeek V3 & 95.3 & 74.3 & 59.1 & 87.1 \\
            Llama 3.3 70b-Instruct & 95.1 & 72.7 & 50.5 & 86.0 \\
            Gemini 1.5 Flash & 90.3 & 69.9 & 39.5 & 78.9 \\
            \hline
        \end{tabular}
    }
    \vspace{-0.1in}
    \caption{Comparison of teacher models' performance across ARC-Challenge, MedMCQA, GPQA, and MMLU benchmarks.}
    \label{tab:performance_gpt4o_teacher}
    \vspace{-0.1in}
\end{table}

\begin{table*}[t]
    \centering
    \resizebox{\textwidth}{!}{ 
        \begin{tabular}{lc|cccc|cccc|cccc}
            \toprule
            \multirow{2}{*}{\textbf{Target SLM}} & \multirow{2}{*}{\textbf{Original}} 
            & \multicolumn{4}{c|}{\textbf{Graph Only}} 
            & \multicolumn{4}{c|}{\textbf{Evidence Only}} 
            & \multicolumn{4}{c}{\textbf{Graph and Evidence Combined}} \\
            \cmidrule(lr){3-6} \cmidrule(lr){7-10} \cmidrule(lr){11-14}
            &  & 5 & 10 & 15 & 20
            & 5 & 10 & 15 & 20
            & 5 & 10 & 15 & 20 \\
            \midrule
            Phi-3.5-mini-instruct & 33.33 & 31.87 & 36.26 & 36.63 & 37.91 & 34.43 & 37.91 & 39.56 & 40.29 & 33.52 & 35.90 & 36.08 & 39.01 \\
            Qwen2.5-3B-Instruct   & 32.23 & 32.97 & 36.45 & 34.07 & 36.81 & 32.60 & 37.55 & 37.55 & 37.55 & 33.70 & 38.10 & 37.91 & 37.91 \\
            Llama-3.2-3B-Instruct & 31.11 & 31.32 & 33.70 & 34.62 & 35.90 & 31.68 & 35.35 & 37.18 & 37.55 & 31.14 & 35.71 & 36.45 & 35.53 \\
            Llama-3.1-8B-Instruct & 32.23 & 37.55 & 38.28 & 36.81 & 38.83 & 38.46 & 40.29 & 39.93 & 40.29 & 37.36 & 39.56 & 40.84 & 40.11 \\
            Qwen2.5-7B-Instruct   & 31.50 & 34.43 & 38.64 & 38.64 & 40.11 & 38.28 & 40.66 & 41.21 & 42.12 & 37.91 & 41.76 & 41.94 & 42.49 \\
            gemma-2-9b-it         & 34.80 & 39.19 & 39.56 & 38.46 & 39.38 & 40.66 & 41.21 & 39.38 & 40.11 & 39.19 & 41.76 & 39.74 & 41.03 \\
            gemma-2-2b-it         & 28.75 & 32.42 & 32.23 & 32.78 & 33.70 & 32.97 & 34.98 & 36.26 & 36.81 & 33.33 & 35.16 & 35.71 & 37.55 \\
            \bottomrule
        \end{tabular}
    }
    \vspace{-0.1in}
    \caption{Comparison of results on the GPQA using GPT-4o as the teacher model.}
    \label{tab:model-performance-gpt-gpqa}
    \vspace{0.1in}

    \centering
    \resizebox{\textwidth}{!}{ 
        \begin{tabular}{lc|cccc|cccc|cccc}
            \toprule
            \multirow{2}{*}{\textbf{Target SLM}} & \multirow{2}{*}{\textbf{Original}}
            & \multicolumn{4}{c|}{\textbf{Graph Only}} 
            & \multicolumn{4}{c|}{\textbf{Evidence Only}} 
            & \multicolumn{4}{c}{\textbf{Graph and Evidence Combined}} \\
            \cmidrule(lr){3-6} \cmidrule(lr){7-10} \cmidrule(lr){11-14}
            &  & 5 & 10 & 15 & 20
            & 5 & 10 & 15 & 20
            & 5 & 10 & 15 & 20 \\
            \midrule
            Phi-3.5-mini-instruct  & 68.7  & 69.88 & 72.39 & 73.78 & 74.82 & 73.53 & 75.80 & 77.21 & 77.80 & 73.18 & 75.82 & 77.05 & 77.46 \\
            Qwen2.5-3B-Instruct    & 65.4  & 65.77 & 68.37 & 69.61 & 70.24 & 69.43 & 71.83 & 73.02 & 73.76 & 69.31 & 71.95 & 72.79 & 73.20 \\
            Llama-3.2-3B-Instruct  & 60.3  & 65.11 & 68.00 & 69.77 & 70.44 & 69.31 & 71.84 & 73.39 & 74.35 & 69.31 & 71.43 & 73.39 & 73.90 \\
            Llama-3.1-8B-Instruct  & 67.9  & 68.77 & 71.32 & 72.70 & 73.63 & 71.93 & 74.62 & 76.11 & 76.93 & 72.11 & 74.72 & 76.10 & 76.77 \\
            Qwen2.5-7B-Instruct    & 71.7  & 71.45 & 73.42 & 74.17 & 75.24 & 73.96 & 76.29 & 76.94 & 77.82 & 74.26 & 76.26 & 76.95 & 77.42 \\
            gemma-2-9b-it          & 71.8  & 72.02 & 73.93 & 74.95 & 75.79 & 74.41 & 76.20 & 77.15 & 77.80 & 74.51 & 76.25 & 77.24 & 77.55 \\
            gemma-2-2b-it          & 56.8  & 62.16 & 64.50 & 66.66 & 67.44 & 66.19 & 68.91 & 70.39 & 71.16 & 66.24 & 68.93 & 70.37 & 70.64 \\
            \bottomrule
        \end{tabular}
    }
    \vspace{-0.1in}
    \caption{Comparison of results on the MMLU using GPT-4o as the teacher model.}
    \label{tab:model-performance-gpt-mmlu}
    \vspace{0.1in}
    
        \centering
    \resizebox{\textwidth}{!}{ 
        \begin{tabular}{lc|cccc|cccc|cccc}
            \toprule
            \multirow{2}{*}{\textbf{Target SLM}} & \multirow{2}{*}{\textbf{Original}}
            & \multicolumn{4}{c|}{\textbf{Graph Only}} 
            & \multicolumn{4}{c|}{\textbf{Evidence Only}} 
            & \multicolumn{4}{c}{\textbf{Graph and Evidence Combined}} \\
            \cmidrule(lr){3-6} \cmidrule(lr){7-10} \cmidrule(lr){11-14}
            &  & 5 & 10 & 15 & 20
            & 5 & 10 & 15 & 20
            & 5 & 10 & 15 & 20 \\
            \midrule
            Phi-3.5-mini-instruct & 55.41 & 59.19 & 62.25 & 65.43 & 67.01 & 64.88 & 68.04 & 68.80 & 68.97 & 63.11 & 67.08 & 68.42 & 68.71 \\
            Qwen2.5-3B-Instruct   & 51.59 & 53.79 & 58.31 & 61.06 & 62.85 & 60.22 & 65.14 & 66.48 & 67.13 & 59.53 & 63.59 & 65.34 & 64.93 \\
            Llama-3.2-3B-Instruct & 50.90 & 58.19 & 61.44 & 64.26 & 65.79 & 64.52 & 67.34 & 67.99 & 68.78 & 63.09 & 66.53 & 67.34 & 68.11 \\
            Llama-3.1-8B-Instruct & 59.14 & 63.52 & 65.29 & 67.34 & 68.71 & 66.60 & 69.04 & 69.52 & 69.81 & 66.12 & 68.42 & 69.02 & 69.57 \\
            Qwen2.5-7B-Instruct   & 56.08 & 58.26 & 61.32 & 64.04 & 65.79 & 63.90 & 67.68 & 68.13 & 68.54 & 64.57 & 67.20 & 67.54 & 68.28 \\
            gemma-2-9b-it         & 56.83 & 62.80 & 65.55 & 66.39 & 67.75 & 66.22 & 68.90 & 69.33 & 68.99 & 66.41 & 68.92 & 69.23 & 69.35 \\
            gemma-2-2b-it         & 42.91 & 52.38 & 58.09 & 60.79 & 62.44 & 58.74 & 64.31 & 65.54 & 66.36 & 58.52 & 63.76 & 65.19 & 65.96 \\
            \bottomrule
        \end{tabular}
    }
    \vspace{-0.1in}
    \caption{Comparison of results on the MedMCQA using Llama 3.3 70B as the teacher model.}
    \label{tab:model-performance-llama-medmcqa}
\end{table*}

\section{Additional Results and Ablations}
\subsection{More Ablation on Teacher Model}
Tables~\ref{tab:model-performance-gpt-med},~\ref{tab:model-performance-llama-medmcqa},~\ref{tab:model-performance-claude-medmcqa},~\ref{tab:model-performance-gemini-medmcqa}, and~\ref{tab:model-performance-deepseek-medmcqa} illustrate the performance of various teacher models on the MedMCQA dataset. The results on MedMCQA demonstrate improvement across all small-scale models when incorporating additional context, whether through graph-based, evidence-based, or combined distillation approaches. 

Under the GPT-4o teacher model, impressive improvements are seen. Phi-3.5-mini-instruct improves from 55.41\% to 74.13\% in the best configuration (20 evidence). Similarly, Qwen2.5-3B-Instruct experienced a 21.23\% improvement, rising from 51.59\% to 72.82\%. Most notably, smaller models show dramatic improvements. For instance, the smaller gemma-2-2b-it, which originally scored 42.91\%, achieves up to 72.44\% in the 20-evidence experiment, representing a 29.52\% increase.

The pattern of stronger performance on smaller student models is consistent across Claude 3.5 Sonnet, Llama 3.3 70B, Gemini 1.5 Flash, and DeepSeek V3 as teachers. For gemma-2-2b-it, performance increases up to 23.05\% for Llama (20 graph and evidence), up to 25.13\% for Claude (20 graph and evidence), up to 14.70\% for Gemini (20 evidence), and up to 23.5\% for DeepSeek (20 evidence). Across the five teacher models, the best-performing augmentation leads to improvements between 14.7\% and 29.5\%.

Performance gains vary based on model size, with larger models like Llama-3.1-8B and gemma-2-9b-it showing relatively less improvement. Thus, while structured knowledge integration is beneficial across all student models, smaller models gain the most from these enhancements due to their initial performance limitations.

\subsection{Comparison Across Various Student Models}
Figure~\ref{methods_ablation_appendix} presents extra results extending Table~\ref{compare_sota}, further examining the impact of retrieval strategies for Llama-3.2-3B-it and Gemma-2-9B-it. 

Consistently across all benchmarks, \textbf{Evidence Only} retrieval achieves the highest accuracy, reinforcing the importance of direct textual evidence in retrieval-augmented learning. The performance gap between evidence-only distillation and graph-only distillation is particularly noticeable in knowledge-intensive tasks such as MedMCQA and ARC-C, where factual precision is crucial. While graph-based retrieval provides some benefits, its improvements remain more limited, especially in benchmarks requiring extensive factual recall. The results suggest that direct evidence retrieval provides a more reliable source of knowledge for model reasoning, whereas graph-based retrieval alone may not be sufficient to bridge factual gaps. Additionally, increasing the number of retrieved items consistently enhances performance, though the gains diminish beyond 15 items. These findings highlight the effectiveness of evidence-based retrieval as the dominant strategy for boosting model performance in multiple-choice QA tasks.

\begin{table*}[t]
        \centering
    \resizebox{\textwidth}{!}{ 
        \begin{tabular}{lc|cccc|cccc|cccc}
            \toprule
            \multirow{2}{*}{\textbf{Target SLM}} & \multirow{2}{*}{\textbf{Original}} 
            & \multicolumn{4}{c|}{\textbf{Graph Only}} 
            & \multicolumn{4}{c|}{\textbf{Evidence Only}} 
            & \multicolumn{4}{c}{\textbf{Graph and Evidence Combined}} \\
            \cmidrule(lr){3-6} \cmidrule(lr){7-10} \cmidrule(lr){11-14}
            &  & 5 & 10 & 15 & 20
            & 5 & 10 & 15 & 20
            & 5 & 10 & 15 & 20 \\
            \midrule
            Phi-3.5-mini-instruct & 55.41 & 60.51 & 64.07 & 65.14 & 66.15 & 65.03 & 68.78 & 69.78 & 70.26 & 66.24 & 70.16 & 70.28 & 70.86 \\
            Qwen2.5-3B-Instruct   & 51.59 & 55.73 & 58.47 & 60.32 & 61.22 & 60.29 & 64.38 & 65.50 & 65.55 & 62.40 & 66.39 & 67.68 & 68.35 \\
            Llama-3.2-3B-Instruct & 50.90 & 60.15 & 62.80 & 64.69 & 64.95 & 64.48 & 67.10 & 68.56 & 68.95 & 65.65 & 68.68 & 69.21 & 70.28 \\
            Llama-3.1-8B-Instruct & 59.14 & 64.62 & 67.37 & 68.95 & 69.69 & 68.47 & 70.74 & 71.74 & 72.17 & 68.85 & 71.69 & 71.69 & 72.20 \\
            Qwen2.5-7B-Instruct   & 56.08 & 61.99 & 63.83 & 65.67 & 66.72 & 65.36 & 68.35 & 70.07 & 70.36 & 66.58 & 69.78 & 70.14 & 70.93 \\
            gemma-2-9b-it         & 56.83 & 64.74 & 67.73 & 69.38 & 70.69 & 67.54 & 70.62 & 71.34 & 71.86 & 69.18 & 71.43 & 71.48 & 71.91 \\
            gemma-2-2b-it         & 42.91 & 52.21 & 56.23 & 57.76 & 58.69 & 58.28 & 62.11 & 63.93 & 65.43 & 61.13 & 66.29 & 66.65 & 68.04 \\
            \bottomrule
        \end{tabular}
    }
    \vspace{-0.1in}
    \caption{Comparison of results on the MedMCQA using Claude 3.5 Sonnet as the teacher model.}
    \label{tab:model-performance-claude-medmcqa}
    \vspace{0.1in}
    
        \centering
    \resizebox{\textwidth}{!}{ 
        \begin{tabular}{lc|cccc|cccc|cccc}
            \toprule
            \multirow{2}{*}{\textbf{Target SLM}} & \multirow{2}{*}{\textbf{Original}} 
            & \multicolumn{4}{c|}{\textbf{Graph Only}} 
            & \multicolumn{4}{c|}{\textbf{Evidence Only}} 
            & \multicolumn{4}{c}{\textbf{Graph and Evidence Combined}} \\
            \cmidrule(lr){3-6} \cmidrule(lr){7-10} \cmidrule(lr){11-14}
            &  & 5 & 10 & 15 & 20
            & 5 & 10 & 15 & 20
            & 5 & 10 & 15 & full \\
            \midrule
            Phi-3.5-mini-instruct & 55.41 & 57.71 & 58.50 & 58.91 & 59.26 & 59.36 & 60.77 & 61.18 & 61.06 & 58.88 & 59.74 & 60.03 & 60.60 \\
            Qwen2.5-3B-Instruct   & 51.59 & 52.88 & 55.30 & 55.08 & 56.16 & 55.22 & 57.06 & 57.66 & 58.24 & 54.63 & 56.63 & 57.04 & 57.66 \\
            Llama-3.2-3B-Instruct & 50.90 & 57.88 & 58.88 & 59.62 & 59.77 & 60.27 & 61.20 & 61.58 & 61.51 & 59.77 & 60.65 & 60.60 & 61.01 \\
            Llama-3.1-8B-Instruct & 59.14 & 61.37 & 61.94 & 62.99 & 63.18 & 62.35 & 63.23 & 63.78 & 64.26 & 61.53 & 62.56 & 63.09 & 63.38 \\
            Qwen2.5-7B-Instruct   & 56.08 & 58.71 & 60.08 & 59.84 & 59.96 & 59.17 & 60.55 & 60.77 & 61.49 & 59.53 & 60.60 & 60.58 & 61.13 \\
            gemma-2-9b-it         & 56.83 & 59.86 & 60.77 & 61.34 & 61.37 & 60.36 & 61.44 & 61.65 & 61.65 & 60.12 & 61.42 & 61.32 & 61.46 \\
            gemma-2-2b-it         & 42.91 & 52.35 & 54.00 & 54.24 & 55.51 & 54.98 & 56.83 & 57.21 & 57.61 & 54.96 & 56.01 & 56.32 & 57.06 \\
            \bottomrule
        \end{tabular}
    }
    \vspace{-0.1in}
    \caption{Comparison of results on the MedMCQA using Gemini 1.5 Flash as the teacher model.}
    \label{tab:model-performance-gemini-medmcqa}
    \vspace{0.1in}
    
    \centering
    \resizebox{\textwidth}{!}{ 
        \begin{tabular}{lc|cccc|cccc|cccc}
            \toprule
            \multirow{2}{*}{\textbf{Target SLM}} & \multirow{2}{*}{\textbf{Original}} 
            & \multicolumn{4}{c|}{\textbf{Graph Only}} 
            & \multicolumn{4}{c|}{\textbf{Evidence Only}} 
            & \multicolumn{4}{c}{\textbf{Graph and Evidence Combined}} \\
            \cmidrule(lr){3-6} \cmidrule(lr){7-10} \cmidrule(lr){11-14}
            &  & 5 & 10 & 15 & 20
            & 5 & 10 & 15 & 20
            & 5 & 10 & 15 & full \\
            \midrule
            Phi-3.5-mini-instruct & 55.41 & 63.90 & 65.00 & 66.12 & 66.34 & 67.42 & 68.71 & 68.95 & 69.47 & 67.27 & 68.16 & 68.49 & 68.83 \\
            Qwen2.5-3B-Instruct   & 51.59 & 59.55 & 62.28 & 63.18 & 63.42 & 65.12 & 65.91 & 66.27 & 66.29 & 64.52 & 65.91 & 66.08 & 66.32 \\
            Llama-3.2-3B-Instruct & 50.90 & 63.11 & 65.05 & 66.56 & 67.01 & 67.61 & 68.66 & 69.16 & 69.14 & 67.27 & 68.56 & 68.80 & 68.73 \\
            Llama-3.1-8B-Instruct & 59.14 & 66.36 & 66.96 & 68.35 & 68.68 & 68.90 & 70.12 & 70.40 & 70.50 & 68.52 & 69.23 & 69.40 & 69.90 \\
            Qwen2.5-7B-Instruct   & 56.08 & 63.59 & 65.19 & 66.56 & 67.06 & 67.89 & 68.49 & 68.76 & 68.83 & 67.65 & 68.28 & 68.54 & 68.95 \\
            gemma-2-9b-it         & 56.83 & 65.98 & 66.65 & 68.06 & 69.04 & 68.66 & 69.42 & 69.59 & 69.35 & 68.75 & 69.42 & 69.42 & 69.33 \\
            gemma-2-2b-it         & 42.91 & 58.93 & 61.13 & 62.20 & 63.47 & 63.85 & 65.22 & 66.24 & 66.41 & 63.93 & 65.05 & 65.46 & 65.50 \\
            \bottomrule
        \end{tabular}
    }
    \vspace{-0.1in}
    \caption{Comparison of results on the MedMCQA using DeepSeek V3 as the teacher model.}
    \label{tab:model-performance-deepseek-medmcqa}
    
\end{table*}

\section{More Results on Open QA Dataset}

Following OpenQA~\cite{lin2018denoising}\footnote{\url{https://github.com/thunlp/OpenQA}.}, we provide more results on WebQuestions dataset\footnote{\url{https://github.com/brmson/dataset-factoid-webquestions}.} in Table~\ref{tab:WebQuestions}. Our framework consistently achieves much better performance than the original baseline method.

\begin{table*}[t]
    \centering
    \resizebox{0.68\textwidth}{!}{
        \begin{tabular}{lcccc}
            \toprule
            \textbf{Model} & \textbf{Original} & \textbf{15 Graph} & \textbf{15 Evidence} & \textbf{ 15 Combined} \\
            \midrule
             Phi-3.5-mini-instruct  & 35.23    & 50.72    & 53.77       & 53.10    \\
             Qwen2.5-3B-Instruct    & 31.61    & 50.16    & 52.98       & 53.82    \\
             Llama-3.2-3B-Instruct  & 28.17    & 49.12    & 53.20       & 53.55    \\
             Llama-3.1-8B-Instruct  & 42.81    & 55.83    & 58.21       & 58.12    \\
             Qwen2.5-7B-Instruct    & 38.24    & 54.79    & 57.60       & 56.21    \\
             gemma-2-9b-it          & 45.71    & 56.22    & 59.15       & 58.26    \\
             gemma-2-2b-it          & 24.51    & 49.69    & 52.10       & 51.72    \\
            \bottomrule
        \end{tabular}
    }
    \vspace{-0.1in}
    \caption{Comparison of results on WebQuestions dataset.}
    \label{tab:WebQuestions}
\end{table*}

\section{More Ablations for Analyzing {\bf $\bf \emph{N}$}}

We provide additional ablation studies using larger $N\!=\!$ 30, 40, and 50 to show the effect and analysis of $N$ in Table~\ref{tab:add_ablation}. Our results indicate that increasing $N$ further does not lead to significant performance improvement. This is intuitive, as only a limited amount of evidence is typically needed to answer a question, excessive input may introduce noise and confuse the SLM in identifying the most relevant information. This finding aligns with our main experiments, which show that performance generally peaks around 15 evidence pieces, while adding more often leads to redundancy and slight performance degradation due to noise.

\begin{table*}[t]
    \centering
    \resizebox{1.0\textwidth}{!}{
        \begin{tabular}{lc|ccc|ccc|ccc}
            \toprule
            \textbf{Model} & \textbf{Original} & \textbf{30 Graph} & \textbf{30 Evidence} & \textbf{ 30 Combined} & \textbf{40 Graph} & \textbf{40 Evidence} & \textbf{ 40 Combined} & \textbf{50 Graph} & \textbf{50 Evidence} & \textbf{ 50 Combined} \\
            \midrule
             Phi-3.5-mini-instruct  & 33.33    & 38.41    & 40.87       & 39.49       & 37.18    & 39.61       & 38.44       & 36.99    & 39.52       & 38.27    \\
             Qwen2.5-3B-Instruct    & 32.23    & 37.12    & 37.29       & 38.41       & 36.55    & 36.88       & 37.11       & 36.30    & 37.08       & 37.47    \\
             Llama-3.2-3B-Instruct  & 31.11    & 35.72    & 37.80       & 35.25       & 35.22    & 37.04       & 34.72       & 35.64    & 37.40       & 35.39    \\
             Llama-3.1-8B-Instruct  & 32.23    & 39.23    & 40.31       & 39.78       & 38.43    & 39.63       & 39.56       & 38.77    & 40.28       & 40.27    \\
             Qwen2.5-7B-Instruct    & 31.50    & 40.42    & 41.90       & 42.41       & 39.73    & 41.45       & 41.74       & 40.38    & 41.58       & 41.01    \\
             gemma-2-9b-it          & 34.80    & 39.79    & 39.79       & 41.29       & 38.94    & 39.41       & 40.61       & 40.07    & 40.84       & 41.77    \\
             gemma-2-2b-it          & 28.75    & 33.46    & 37.10       & 37.33       & 33.08    & 36.33       & 36.78       & 34.50    & 35.64       & 36.43    \\
             \bottomrule
         \end{tabular}
    }
    \vspace{-0.1in}
    \caption{Additional ablation studies using larger $N\!=\!$ 30, 40, and 50 for the effect and analysis of $N$.}
    \label{tab:add_ablation}
\end{table*}

\section{Upper Bound of Teachers' Performance}
Table \ref{tab:performance_gpt4o_teacher} presents a comparison across different teacher models. The results demonstrate that our distillation framework enables student models to achieve performance remarkably close to this upper bound. On ARC-C and MedMCQA, the best student models, such as Gemma-2-9b-it and Llama-3.1-8B-Instruct, perform within 2.0–2.5\% of GPT-4o, demonstrating minimal performance loss after distillation. On GPQA and MMLU, student models show competitive performance, with Gemma-2-9b-it reaching 77.8\% on MMLU, demonstrating that our approach narrows the gap between student and teacher models. While GPT-4o represents the upper bound for performance, our results show that smaller models can reach highly competitive accuracy levels.

\section{Prompt Used to Produce Evidence}

The following system prompt and user prompt are used for evidence generation:

\noindent\textbf{System Prompt:} 

\begin{quote}
\em
"You are an assistant in charge of generating factual evidence statements that aid in solving the provided question. Provide only the evidence statements with no additional remarks. Do not give the answer away directly."
\end{quote}

\noindent\textbf{User Prompt:} 

\begin{quote}
\em
"Generate $N$ evidences that pertain to answering the following question: \{q\}"
\end{quote}

This process is repeated for every task within each of the tested benchmarks.

\section{Details of Privacy Sensitive Benchmark Construction}\label{privacy_algo}

To demonstrate the effectiveness of our \name{} framework in preventing the leakage of sensitive information while maintaining answer quality, we constructed a benchmark based on the MMLU dataset. Our approach begins with a diverse sample of 5,000 questions randomly drawn from MMLU. To simulate realistic privacy risks such as those involving the unintentional exposure of personally identifiable information (PII), we use GPT-4o to augment these questions by injecting synthetic yet realistic PII (e.g., fabricated names, email addresses, and affiliations). To mitigate these risks, we use SLM to detect and remove any sensitive information. This redaction process preserves the original semantic content of each query while ensuring that only privacy-safe inputs are forwarded to our \name{} framework. The cleaned questions are then processed by \name{}, which retrieves relevant external knowledge before generating high-quality answers.
The overall process is summarized as follows:
\begin{enumerate}[label=\textbf{Step \arabic*:}, leftmargin=1.25cm]
    \item \textbf{Sampling:} Randomly select 5,000 questions from the MMLU dataset.
    \item \textbf{Augmentation:} Utilize GPT-4o to inject synthetic PII into the selected questions, thereby simulating potential privacy risks.
    \item \textbf{Privacy Filtering:} Apply the SLM to redact the injected PII while preserving the original meaning of the queries.
    \item \textbf{DRAG Processing:} Process the filtered queries through the \name{} framework, which retrieves relevant evidence and generates accurate answers.
\end{enumerate}
This framework prevents sensitive information from being shared while still producing useful and accurate evidence.

\section{Simple Graph Construction} \label{simple_graph}
A simple graph is defined as a graph without multiple edges between any two given nodes, and a multiple graph is defined as a graph that is allowed to contain multiple edges between two nodes \citep{wolframalpha}. In the context of graphs in \name{}, nodes correspond to entities and edges correspond to relationships. Essentially, in the {\em simple graph aggregation approach}, for any two entities $a, b$ that contain multiple relationships, the edges are combined into one aggregated relationship.

For instance, in Figure~\ref{fig:methodology}, in the multigraph, there are two relationships with the entity pair (nymphs, adults). After applying the {\em simple graph aggregation approach}, the two relationships between (nymphs, adults) are consolidated into one relationship. This aggregation is performed by prompting the teacher model with the following prompt: 
\begin{quote}
\em
"You are an assistant in charge of combining the provided statements into one summarized statement. Be concise without losing any of the information."
\end{quote}

We observe that this operation further reduces some degree of redundancy in graph representation and slightly enhances framework efficiency without impacting performance, while it is essentially optional as the improvement is marginal.

\section{Full Pipeline Example}

\newpage

\begin{originalquestion}
Which term best describes the life cycle of an insect that reaches the adult stage without being a pupa?  

\begin{enumerate}
    \item[a.] Incomplete metamorphosis
    \item[b.] Complete metamorphosis
    \item[c.] Alternation of generations
    \item[d.] Spontaneous mutation
\end{enumerate}
\end{originalquestion}

\begin{evidencegen}
\begin{itemize}[leftmargin=0pt]
    \item Incomplete metamorphosis, also known as hemimetabolism, is a type of insect development where the immature stages (nymphs) resemble the adult but lack fully developed wings and reproductive structures.
    \item Insects undergoing incomplete metamorphosis do not have a pupal stage; they transition directly from nymphs to adults.
    \item The nymphs gradually grow larger, molting multiple times until they become adults, which is a hallmark of incomplete metamorphosis.
    \item Common insects like grasshoppers, crickets, cockroaches, and dragonflies exhibit incomplete metamorphosis, all bypassing the pupal stage.
    \item Complete metamorphosis (holometabolism) involves four distinct stages: egg, larva, pupa, and adult. The presence of a pupal stage clearly differentiates it from incomplete metamorphosis.
\end{itemize}
\end{evidencegen}

\begin{evidencefilt}
\begin{itemize}[leftmargin=0pt]
    \item The nymphs gradually grow larger, molting multiple times until they become adults, which is a hallmark of incomplete metamorphosis.
    \item Incomplete metamorphosis, also known as hemimetabolism, is a type of insect development where the immature stages (nymphs) resemble the adult but lack fully developed wings and reproductive structures.
    \item Insects undergoing incomplete metamorphosis do not have a pupal stage; they transition directly from nymphs to adults.
\end{itemize}
\end{evidencefilt}

\begin{ragrel}
\begin{itemize}[leftmargin=0pt]
    \item \textbf{Entity Pair:} (nymphs, adults)  
          \textbf{Relationship:} "Nymphs transition directly into adults, and resemble the adult stage."
    \item \textbf{Entity Pair:} (nymphs, incomplete metamorphosis)  
          \textbf{Relationship:} "Nymphs are a characteristic of incomplete metamorphosis."
    \item \textbf{Entity Pair:} (Incomplete metamorphosis, hemimetabolism)  
          \textbf{Relationship:} "Incomplete metamorphosis is also known as hemimetabolism."
\end{itemize}
\end{ragrel}

\begin{ragfilt}
\begin{itemize}[leftmargin=0pt]
    \item \textbf{Entity Pair:} (nymphs, adults)  
          \textbf{Relationship:} "Nymphs transition directly into adults, and resemble the adult stage."
    \item \textbf{Entity Pair:} (nymphs, incomplete metamorphosis)  
          \textbf{Relationship:} "Nymphs are a characteristic of incomplete metamorphosis."
\end{itemize}
\end{ragfilt}

\begin{responsecol}
\begin{itemize}[leftmargin=0pt]
    \item \textbf{Answer with Evidence and Graph:} Incomplete metamorphosis (a)  
    \item \textbf{Answer with no Context:} Complete metamorphosis (b) 
\end{itemize}
\end{responsecol}

\begin{responseeval}
\begin{itemize}[leftmargin=0pt]
    \item \textbf{Evaluation of Answer with Evidence and Graph:}(a) is correct.  
    \item \textbf{Evaluation of Answer with no Context:} (b) is incorrect.
\end{itemize}

\end{responseeval}

\subsection{Privacy Benchmark Sample Questions}
\subsection*{Example 1}

\begin{originalquestion}
Which of the following amino acids cannot provide a substrate for gluconeogenesis?
\end{originalquestion}

\begin{privacyissuequestion}
Which of the following amino acids cannot provide a substrate for gluconeogenesis? Dr. Amelia Thompson, who wrote an article on amino acid metabolism for Biochem Review, can be contacted at athompson93@biochemjournal.fake.
\end{privacyissuequestion}

\begin{finalquestion}
What amino acids cannot provide a substrate for gluconeogenesis? (edited)

\begin{enumerate}[label=\alph*.]
\item Leucine and lysine
\item Alanine and glycine
\item Aspartate and glutamate
\item Serine and threonine
\end{enumerate}
\end{finalquestion}

\begin{evidencegen}
\begin{itemize}[leftmargin=0pt]
\item Gluconeogenesis is the metabolic pathway through which glucose is synthesized from non-carbohydrate sources, including most amino acids.
\item Amino acids can be classified as glucogenic, ketogenic, or both, based on their metabolic fate.
\item Leucine and lysine are the only exclusively ketogenic amino acids, meaning they cannot serve as substrates for gluconeogenesis.
\item Ketogenic amino acids are metabolized into acetyl-CoA or acetoacetate, which cannot be converted into glucose.
\item Glucogenic amino acids, such as alanine and glycine, can be converted into intermediates of the citric acid cycle, ultimately leading to glucose production.
\end{itemize}
\end{evidencegen}

\begin{evidencefilt}
\begin{itemize}[leftmargin=0pt]
\item Leucine and lysine are the only exclusively ketogenic amino acids, meaning they cannot serve as substrates for gluconeogenesis.
\item Ketogenic amino acids are metabolized into acetyl-CoA or acetoacetate, which cannot be converted into glucose.
\end{itemize}
\end{evidencefilt}

\begin{ragrel}
\begin{itemize}[leftmargin=0pt]
\item \textbf{Entity Pair:} (Leucine, gluconeogenesis)\\
\textbf{Relationship:} "Leucine cannot be used as a substrate for gluconeogenesis."
\item \textbf{Entity Pair:} (Lysine, gluconeogenesis)\\
\textbf{Relationship:} "Lysine cannot be used as a substrate for gluconeogenesis."
\item \textbf{Entity Pair:} (Leucine, lysine)\\
\textbf{Relationship:} "Leucine and lysine are both exclusively ketogenic amino acids."
\end{itemize}
\end{ragrel}

\begin{ragfilt}
\begin{itemize}[leftmargin=0pt]
\item \textbf{Entity Pair:} (Leucine, gluconeogenesis)
\textbf{Relationship:} "Leucine cannot be used as a substrate for gluconeogenesis."
\item \textbf{Entity Pair:} (Lysine, gluconeogenesis)
\textbf{Relationship:} "Lysine cannot be used as a substrate for gluconeogenesis."
\end{itemize}
\end{ragfilt}

\begin{responsecol}
\begin{itemize}[leftmargin=0pt]
\item \textbf{Answer with Evidence and Graph:} Leucine and lysine (a)
\item \textbf{Answer with no Context:} Leucine and lysine (a)
\end{itemize}
\end{responsecol}

\begin{responseeval}
\begin{itemize}[leftmargin=0pt]
\item \textbf{Evaluation of Answer with Evidence and Graph:} (a) is correct.
\item \textbf{Evaluation of Answer with no Context:} (a) is correct.
\end{itemize}
\end{responseeval}

\subsection*{Example 2}
\begin{originalquestion} 
Another name for the fallacy of false dilemma is
\end{originalquestion}

\begin{privacyissuequestion}
Another name for the fallacy of false dilemma, which Dr. Emily Winters, who teaches logic at Crestwood University (emily.winters@crestwood.edu), often references in her lectures, is
\end{privacyissuequestion}

\begin{finalquestion}
What is another name for the fallacy of false dilemma?

\begin{enumerate}
    \item[a.] False dichotomy
    \item[b.] Slippery slope
    \item[c.] Circular reasoning
    \item[d.] Hasty generalization
\end{enumerate}
\end{finalquestion}

\begin{evidencegen}
\textbf{Evidence Generation}
\begin{itemize}[leftmargin=0pt]
    \item The false dilemma fallacy, also known as a false dichotomy, occurs when a situation is presented as having only two alternatives when more options exist.  
    \item This fallacy is often used in arguments to force a choice between two extremes, ignoring potential middle ground or alternative perspectives.  
    \item The false dichotomy fallacy misrepresents the complexity of an issue by reducing it to a binary decision.  
    \item Other logical fallacies, such as slippery slope or hasty generalization, involve different reasoning errors but do not equate to a false dilemma.  
    \item Circular reasoning involves using the conclusion as one of the premises, which is distinct from the structure of a false dilemma.  
\end{itemize}
\end{evidencegen}

\begin{evidencefilt}
\begin{itemize}[leftmargin=0pt]
    \item The false dilemma fallacy, also known as a false dichotomy, occurs when a situation is presented as having only two alternatives when more options exist.  
    \item This fallacy is often used in arguments to force a choice between two extremes, ignoring potential middle ground or alternative perspectives.  
    \item The false dichotomy fallacy misrepresents the complexity of an issue by reducing it to a binary decision.  
\end{itemize}
\end{evidencefilt}

\begin{ragrel}
\begin{itemize}[leftmargin=0pt]
    \item \textbf{Entity Pair:} (false dilemma, false dichotomy)  
          \textbf{Relationship:} "False dilemma is also known as false dichotomy."  
    \item \textbf{Entity Pair:} (false dichotomy, binary decision)  
          \textbf{Relationship:} "A false dichotomy incorrectly reduces a complex issue to a binary decision."  
    \item \textbf{Entity Pair:} (false dilemma, extreme choices)  
          \textbf{Relationship:} "False dilemma forces a choice between two extremes, ignoring other options."  
\end{itemize}
\end{ragrel}

\begin{ragfilt}
\begin{itemize}[leftmargin=0pt]
    \item \textbf{Entity Pair:} (false dilemma, false dichotomy)  
          \textbf{Relationship:} "False dilemma is also known as false dichotomy."  
    \item \textbf{Entity Pair:} (false dichotomy, binary decision)  
          \textbf{Relationship:} "A false dichotomy incorrectly reduces a complex issue to a binary decision."  
\end{itemize}
\end{ragfilt}

\begin{responsecol}
\begin{itemize}[leftmargin=0pt]
    \item \textbf{Answer with Evidence and Graph:} False dichotomy (a)  
    \item \textbf{Answer with no Context:} Slippery slope (b)  
\end{itemize}
\end{responsecol}

\begin{responseeval}
\begin{itemize}[leftmargin=0pt]
    \item \textbf{Evaluation of Answer with Evidence and Graph:} (a) is correct.  
    \item \textbf{Evaluation of Answer with no Context:} (b) is incorrect.  
\end{itemize}
\end{responseeval}

\subsection*{Example 3}
\begin{originalquestion}
Each resonance form of the nitrate ion, NO$_3^-$, has how many sigma and how many pi bonds?    
\end{originalquestion}

\begin{privacyissuequestion}
\textbf{Modified Question with Privacy Issue:}  
Each resonance form of the nitrate ion, NO$_3^-$, has how many sigma and how many pi bonds? Dr. Emily Greene, who resides at 123 Chemistry Lane, Springfield, and can be contacted at \texttt{emily.greene@chemresearch.org}, explored this topic in her recent publication.
\end{privacyissuequestion}

\begin{finalquestion}  
What is the number of sigma and pi bonds in each resonance form of the nitrate ion, NO$_3^-$?

\begin{enumerate}
    \item[a.] 3 sigma bonds, 1 pi bond
    \item[b.] 4 sigma bonds, 2 pi bonds
    \item[c.] 3 sigma bonds, 2 pi bonds
    \item[d.] 5 sigma bonds, 1 pi bond
\end{enumerate}
\end{finalquestion}

\begin{evidencegen}
\begin{itemize}[leftmargin=0pt]
    \item The nitrate ion (NO$_3^-$) has three resonance structures, each with one nitrogen-oxygen double bond and two nitrogen-oxygen single bonds.  
    \item Each nitrogen-oxygen double bond contains one sigma bond and one pi bond.  
    \item The nitrogen-oxygen single bonds contain one sigma bond each.  
    \item In total, each resonance form of NO$_3^-$ contains 3 sigma bonds from the single bonds and 1 pi bond from the double bond.
\end{itemize}
\end{evidencegen}

\begin{evidencefilt}
\begin{itemize}[leftmargin=0pt]
    \item The nitrate ion (NO$_3^-$) has three resonance structures, each with one nitrogen-oxygen double bond and two nitrogen-oxygen single bonds.  
    \item Each nitrogen-oxygen double bond contains one sigma bond and one pi bond.  
    \item The nitrogen-oxygen single bonds contain one sigma bond each.  
\end{itemize}    
\end{evidencefilt}

\begin{ragrel}
\begin{itemize}[leftmargin=0pt]
    \item \textbf{Entity Pair:} (nitrate ion, resonance structures)  
          \textbf{Relationship:} "The nitrate ion has three resonance structures."  
    \item \textbf{Entity Pair:} (double bond, sigma bond)  
          \textbf{Relationship:} "A nitrogen-oxygen double bond contains one sigma bond."  
    \item \textbf{Entity Pair:} (double bond, pi bond)  
          \textbf{Relationship:} "A nitrogen-oxygen double bond contains one pi bond."  
    \item \textbf{Entity Pair:} (single bond, sigma bond)  
          \textbf{Relationship:} "Each nitrogen-oxygen single bond contains one sigma bond."  
\end{itemize}
\end{ragrel}

\begin{ragfilt}
\begin{itemize}[leftmargin=0pt]
    \item \textbf{Entity Pair:} (nitrate ion, resonance structures)  
          \textbf{Relationship:} "The nitrate ion has three resonance structures."  
    \item \textbf{Entity Pair:} (double bond, sigma bond)  
          \textbf{Relationship:} "A nitrogen-oxygen double bond contains one sigma bond."  
    \item \textbf{Entity Pair:} (double bond, pi bond)  
          \textbf{Relationship:} "A nitrogen-oxygen double bond contains one pi bond."  
\end{itemize}    
\end{ragfilt}

\begin{responsecol}
\begin{itemize}[leftmargin=0pt]
    \item \textbf{Answer with Evidence and Graph:} 3 sigma bonds, 1 pi bond (a)  
    \item \textbf{Answer with no Context:} 4 sigma bonds, 2 pi bonds (b)  
\end{itemize}
\end{responsecol}

\begin{responseeval}
\begin{itemize}[leftmargin=0pt]
    \item \textbf{Evaluation of Answer with Evidence and Graph:} (a) is correct.  
    \item \textbf{Evaluation of Answer with no Context:} (b) is incorrect.  
\end{itemize}    
\end{responseeval}

\end{document}